\documentclass{article} 
\usepackage{iclr2027_conference,times}

\usepackage{amsmath,amsfonts,bm}

\def\eqref#1{equation~\ref{#1}}

\def\1{\bm{1}}

\DeclareMathAlphabet{\mathsfit}{\encodingdefault}{\sfdefault}{m}{sl}
\SetMathAlphabet{\mathsfit}{bold}{\encodingdefault}{\sfdefault}{bx}{n}

\newcommand{\E}{\mathbb{E}}

\newcommand{\R}{\mathbb{R}}

\newcommand{\Cov}{\mathrm{Cov}}

\DeclareMathOperator{\Tr}{Tr}

\renewcommand{\eqref}[1]{\textup{(\ref{#1})}}
\usepackage{amsmath,amssymb,amsthm}  
\usepackage{bm}                      
\usepackage{array}                   
\usepackage{booktabs,multirow}       
\usepackage{enumitem}                
\usepackage{xcolor}                  
\usepackage{graphicx}                
\usepackage{wrapfig}
\usepackage{placeins}               
\usepackage[skip=0pt,labelfont=bf]{caption} 
\usepackage{hyperref}
\usepackage{cleveref}
\makeatletter
\newcommand{\appendixcontents}{\@starttoc{toc}}  
\makeatother

\definecolor{flodarkpurple}{rgb}{0.288,0.1196,0.7}
\definecolor{teal}{rgb}{0.07450980392156863, 0.5137254901960784, 0.2784313725490196}
\definecolor{blue}{rgb}{0.047, 0.365, 0.647}
\newcommand{\rebuttal}[1]{#1}
\definecolor{rebuttalblue}{HTML}{000000}
\hypersetup{
  colorlinks=true,
  linkcolor=flodarkpurple,
  citecolor=teal,
  urlcolor=blue
}

\newcounter{algo}
\newenvironment{algo}[1]{%
  \begin{figure}[t]
  \refstepcounter{algo}%
  \hrule height 0.9pt \vspace{3pt}
  \noindent\textbf{Algorithm \thealgo} #1\\[1pt]
  \hrule height 0.4pt \vspace{4pt}
  \small
}{%
  \vspace{3pt} \hrule height 0.9pt
  \end{figure}
}
\newenvironment{algosteps}{%
  \begin{enumerate}[leftmargin=2.1em, itemsep=1.5pt, topsep=3pt, parsep=0pt, label=\arabic*:]
}{%
  \end{enumerate}
}

\newtheorem{theorem}{Theorem}
\newtheorem{lemma}{Lemma}
\newtheorem{proposition}{Proposition}
\newtheorem{corollary}{Corollary}
\theoremstyle{remark}

\renewcommand{\Tr}{\operatorname{tr}} 
\newcommand{\Normal}{\mathcal{N}}
\newcommand{\D}{\mathcal{D}}
\newcommand{\thetaMAP}{\theta_{\mathrm{MAP}}}
\newcommand{\thetaMAPbar}{\bar{\theta}_{\mathrm{MAP}}} 
\newcommand{\thetaL}{\theta}
\newcommand{\diag}{\operatorname{diag}}
\newcommand{\dd}{\mathrm{d}}
\newcommand{\transpose}{\mathsf{T}}
\newcommand{\preprintnotice}[1]{\begingroup\renewcommand{\thefootnote}{}\footnotetext{#1}\endgroup}
\makeatletter
\newcommand{\quietthanks}[1]{\begingroup\def\@makefnmark{}\thanks{#1}\endgroup}  
\makeatother

\title{Coherent Moment Propagation: Epistemic \\ Uncertainty for Pre-trained Robot Policies}

\author{\parbox{\dimexpr\textwidth-2\tabcolsep}{\centering\vspace{8pt}
  Miguel Saavedra-Ruiz$^{1,2,3,*}$\quietthanks{Work done during an internship at NVIDIA.}, Anqi Li$^{1}$, Claudia D'Arpino$^{1}$,\\  Alperen Degirmenci$^{1}$, Tucker Hermans$^{1,4}$ and Fabio Ramos$^{1,5}$\\[5pt]
  \normalfont $^{1}$NVIDIA \quad $^{2}$Universit\'e de Montr\'eal \quad $^{3}$Mila -- Quebec AI Institute\\
  \normalfont $^{4}$University of Utah \quad $^{5}$University of Sydney\\[5pt]
  \normalfont Project page: \url{https://comp-uq.github.io/comp-uq}}%
}

\iclrfinalcopy 

\begin{document}
\maketitle
\preprintnotice{Preprint. Under review.}
\addtocontents{toc}{\protect\setcounter{tocdepth}{-10}}  
\vspace{-11pt}  
\lhead{}\renewcommand{\headrulewidth}{0pt}

\begin{abstract}
Flow matching policies underpin the state-of-the-art in robot learning, with impressive results ranging from long-horizon planning to dexterous manipulation. However, they provide no measure of whether their predicted commands can be trusted, particularly in novel situations outside their training set. Epistemic uncertainty reflects uncertainty about the model's weights and quantifies this ignorance, but calculating it at inference time from a flow matching policy is not trivial. Actions are produced by an ODE solver, requiring uncertainty to be propagated through the entire integration process. BayesDiff addresses this problem for diffusion models by propagating the mean and variance of the state through the denoising chain under a last-layer Laplace posterior approximation, but it severely underestimates the epistemic uncertainty. In this paper we: i) prove that, for flow matching and diffusion ODEs, BayesDiff's total variance decays at rate $O(1/T)$ as the number of Euler steps $T$ used by the solver increases; ii) introduce Coherent Moment Propagation (CoMP), two recursions inspired by the extended and unscented Kalman filters that fix this problem by carrying the cross-covariance between state and weights; and iii) widen the last-layer posterior with observation embeddings the policy already computes, improving uncertainty estimates without retraining for CoMP and several test-time estimators alike. We demonstrate these benefits on three manipulation benchmarks with applications in failure detection and demonstration selection for fine-tuning, where CoMP corrects BayesDiff's variance collapse while matching or outperforming several uncertainty estimators.
\end{abstract}

\section{Introduction}

Flow matching~\citep{lipman2023flow, liu2023flow, albergo2025stochastic} has become an integral component of robot stacks, powering action heads in state-of-the-art robot policies~\citep{black2024pi0, bjorck2025gr00t, shukor2025smolvla}.  Despite their widespread success, these models lack an inherent mechanism to assess the reliability of their predictions. Epistemic uncertainty provides one such indicator, measuring how much predictions vary across parameter settings consistent with the training data~\citep{kendall2017uncertainties}. However, obtaining this uncertainty at test time is increasingly important given the resources invested in foundation policies~\citep{bjorck2025gr00t}, whose reliability on unfamiliar tasks remains difficult to assess~\citep{xu2025faildetect, gu2025safe}.

Uncertainty quantification of a flow matching-based policy is not trivial, as it requires propagating model uncertainty through the ODE solver that generates robot commands. A common technique to quantify uncertainty is to compare the predictions from an ensemble of policies, whether trained~\citep{lakshminarayanan2017deep, pearce2020uncertainty, berry2024shedding, romer2026save}, generated by a hypernetwork~\citep{chan2024estimating} or drawn from a posterior fit after training~\citep{jazbec2025generative, huang2026deua}. However, training and storing independent model copies is costly at the scale of foundation policies, motivating estimators that work from a single pre-trained policy.

\begin{figure}[t]
  \centering
  \includegraphics[width=\textwidth]{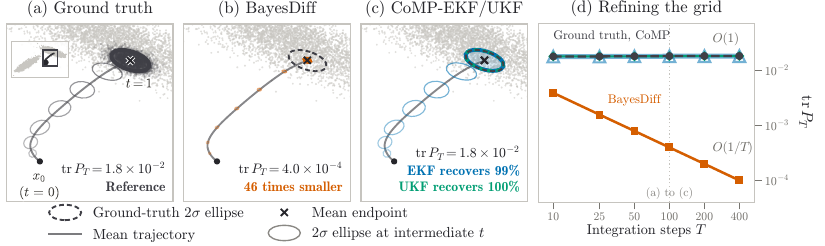}
  \caption{\textbf{Variance collapse in moment propagation.} State covariance $P_T$ at the end of integration of a flow matching model on a two-mode mixture under a
  last-layer Laplace posterior for one sample $x_0$ at
  $T = 100$ Euler steps. (a) The reference is the empirical covariance of
  $4000$ trajectories, one per posterior draw
  $\thetaL^{(m)} \sim \Normal(\thetaMAP, \Sigma_\theta)$. (b) BayesDiff returns $1/46$ of it, consistent with Theorem~\ref{thm:decay}. (c) Our estimators CoMP-EKF and CoMP-UKF carry the cross-covariance between the state and the weights and recover the
  reference to within $1\%$.
  (d) The total variance $\Tr P_T$ at the same seed for every estimator re-run on each grid from $T = 10$
  to $400$ Euler steps. BayesDiff halves with every doubling of $T$, while our methods track the reference regardless of the number of Euler steps.}
  \label{fig:shrinkage}
\end{figure}

Post-hoc Bayesian inference offers a way to obtain epistemic uncertainty from pre-trained models~\citep{daxberger2021laplace, jazbec2025generative}. Particularly, BayesDiff~\citep{kou2024bayesdiff} combines a last-layer Laplace posterior with \emph{moment propagation}, carrying the mean and variance of the denoised state through the sampling chain of a diffusion model~\citep{ho2020ddpm}, with extensions beyond the last layer~\citep{gupta2026quantifying}, to flow matching models~\citep{han2026flow}, and to multi-agent trajectory modeling~\citep{capellera2025unified}. However, we prove that BayesDiff exhibits \emph{variance collapse} on flow matching and diffusion ODEs: under Lipschitz regularity and bounded predictive variance, its estimated total variance is upper bounded by $O(1/T)$ in the number of Euler steps $T$, so it vanishes as the solver is refined; see Figure~\ref{fig:shrinkage} (b) and (d). This creates an artificial tradeoff: finer integration better approximates the ODE solution, yet drives the reported uncertainty to zero.\looseness-1   

To address this limitation, we introduce \emph{\textbf{Co}herent \textbf{M}oment \textbf{P}ropagation} (CoMP), a method that carries the cross-covariance between state and weights, yielding an upper bound on total variance independent of $T$, see Figure~\ref{fig:shrinkage} (c) and (d). We show that BayesDiff's variance collapse results from treating the weights as independent of the state they generate, and derive two recursions inspired by the extended and unscented Kalman filters. Furthermore, to improve uncertainty estimates from a last-layer Laplace approximation~\citep{deng2022ella, calvo2026richer}, we draw inspiration from LULA~\citep{kristiadi2021lula} and widen the last layer with observation embeddings already computed by foundation robot policies, without re-training or changing their action predictions.\looseness-1

In summary, our contributions are:
 
\begin{enumerate}[leftmargin=*, itemsep=0pt, topsep=0pt]
  \item We prove BayesDiff's variance collapse and show its total variance is upper bounded by $O(1/T)$ for flow matching and diffusion ODEs with the number of Euler integration steps $T$;
  \item We derive two new uncertainty estimators, \textbf{CoMP-EKF} and \textbf{CoMP-UKF}, from a linearized sampler recursion that carries state-weight cross-covariance, with a constant total variance bound; and\looseness-1
  \item We improve last-layer Laplace uncertainty estimates by widening the head with existing observation embeddings, without re-training the policy or changing its action predictions.
\end{enumerate}

We demonstrate the practical utility of our uncertainty estimators in the problems of runtime failure detection~\citep{romer2025fiper,xu2025faildetect} and demonstration selection~\citep{agia2025cupid,sirigiri2026diversity} for fine-tuning~\citep{bagatella2025active,romer2026save}. Across three robot policies and three manipulation benchmarks (robomimic~\citep{mandlekar2021robomimic}, LIBERO~\citep{liu2023libero}, and RoboLab~\citep{yang2026robolab}), CoMP is better calibrated than BayesDiff and matches or outperforms several test-time uncertainty estimators for pre-trained robot policies. Lastly, we show that the augmented head further improves uncertainty estimates from CoMP and several baselines.

\section{Preliminaries}
\label{sec:setup}

\paragraph{Flow matching.}
Let $x \in \R^{D}$ denote the state and $u_\theta(x, t)$ a velocity field with parameters $\theta$ that transports noise to data through an ODE over $t \in [0,1]$ with $x_0 \sim \Normal(0, I)$ and $x_1$ drawn from the data distribution. Following the conditional flow matching formulation~\citep{lipman2023flow, albergo2025stochastic}, $u_\theta$ is trained on the affine interpolant $x_t = (1-t)\,x_0 + t\,x_1$ by minimizing: 
\begin{equation}
\label{eq:cfm_loss}
\mathcal{L}_{\mathrm{CFM}}(\theta) =
\E_{t, x_0, x_1}\!\left[\left\|u_\theta(x_t, t) - (x_1 - x_0)\right\|^2\right],
\end{equation}
\noindent
with $t \sim \mathcal{U}[0, 1]$. Sampling integrates $\dd x / \dd t = u_\theta(x, t)$ from a noise sample $x_0 \sim \Normal(0, I)$, using $T$ Euler steps of size $\Delta t\triangleq 1/T$, with $t_i\triangleq i\Delta t$, so that the generated sample $x_{T}$ results from iterating:
\begin{equation}
\label{eq:euler}
x_{i+1} = x_i + \Delta t\, u_\theta(x_i, t_i).
\end{equation}

\paragraph{Flow-based robot policies.}
A vision-language-action (VLA) policy $\pi_\theta(a_{1:H} \mid o)$ maps an observation $o = (s, I, \ell)$ comprising proprioceptive state $s$, camera images $I$ and a language instruction $\ell$ to a distribution over the next $H$ actions~\citep{black2024pi0, bjorck2025gr00t, shukor2025smolvla}. For flow matching-based VLAs, a vision-language backbone encodes $o$ once into $c(o)$, held fixed while an action expert uses~\eqref{eq:euler} to integrate $u_\theta(x, t, c(o))$ from noise to an action chunk:
\begin{equation}
\label{eq:chunk}
x_T \;=\; a_{1:H} \triangleq \left(a_1, \dots, a_H\right) \in \R^{H \times d_a},
\end{equation}
where $H$ is the predicted horizon and $d_a$ the dimension of one action, e.g., an end-effector displacement. The state of the flow is therefore the chunk flattened to $D = H\,d_a$, and the policy executes its first $h \le H$ actions before replanning. In the following we omit the conditioning on $o$.

\paragraph{Laplace approximation.}
The Laplace approximation~\citep{daxberger2021laplace} models the weight posterior $p(\theta \mid \D) \propto p(\D \mid \theta)\,p(\theta)$ of a network trained with data $\D$ as a Gaussian. This approximation allows us to quantify epistemic uncertainty, which reflects variation in the network's predictions for the same input across weights drawn from the posterior. The Gaussian is centered at a posterior mode, with covariance given by the inverse curvature of the negative log-posterior:
\begin{equation}
\label{eq:llla}
p(\theta \mid \D) \approx \Normal\!\left(\theta;\, \theta_{\mathrm{MAP}},\, \Sigma_\theta\right),
\qquad
\Sigma_\theta \triangleq (\left.\nabla^2_\theta \mathcal{L}(\D; \theta)\right|_{\theta_{\mathrm{MAP}}})^{-1},
\end{equation}
\noindent
where the objective $\mathcal{L}(\D; \theta) \triangleq \mathcal{L}_{\mathrm{CFM}}(\D; \theta) + (\lambda/2)\|\theta\|^{2}$ is a negative log-posterior, with the flow matching loss~\eqref{eq:cfm_loss} as the negative log-likelihood $-\log p(\D \mid \theta)$ and weight decay as the negative log-prior under $p(\theta) = \Normal(0, \lambda^{-1} I)$. Computing the full Hessian is costly, so approximations restrict the posterior to the \textbf{last layer}, whose curvature can be estimated from a data subset $\D_{\mathrm{fit}}$, freezing the remaining weights at their MAP~\citep{kristiadi2020llla}. Here, $\thetaL$ denotes the last-layer weights, $\thetaMAP$ their pre-trained value, and $\delta\theta \triangleq \thetaL - \thetaMAP \sim \Normal(0, \Sigma_\theta)$ the deviation of a single posterior draw~\eqref{eq:llla}.\looseness-1

\paragraph{Augmented last-layer posterior.}
A posterior on the last layer can only capture what its input features $\phi(x,t)$ retain. A last layer trained to predict a velocity can over-compress observations, keeping only what the velocity needs and discarding the rest, potentially leading to feature collapse~\citep{vanamersfoort2021feature, mukhoti2023ddu}.  LULA~\citep{kristiadi2021lula} addresses this by augmenting the features feeding the last layer to improve uncertainty calibration, while zero output weights on the added units preserve the network's predictions. For a linear last layer, this gives\looseness-1 
\begin{equation}
\label{eq:cond_head}
u_\theta(x, t) = \theta\, \psi(x, t),
\qquad
\psi(x, t) \triangleq \big(\phi(x, t),\; \zeta\big),
\qquad
\rebuttal{\thetaMAPbar \triangleq} \big(\thetaMAP,\, 0\big),
\end{equation}
\noindent
where $\zeta$ contains the activations of the added hidden units, and $\thetaMAPbar$ denotes the pre-trained weights with zero output weights appended. \rebuttal{In LULA~\citep{kristiadi2021lula}, $\thetaMAPbar$ is chosen so that the policy output is unchanged, but note that it is not necessarily a mode of the augmented posterior. The approximation in~\eqref{eq:llla} with $\thetaMAPbar$ then becomes a curvature-based Gaussian centered at this frozen predictor, with covariance estimated from the augmented loss curvature (Appendix~\ref{app:cond_head}).} However, LULA learns these added features $\zeta$ using an out-of-distribution set, which a pre-trained policy does not provide.\looseness-1

\paragraph{The BayesDiff recursion.}
BayesDiff~\citep{kou2024bayesdiff} introduced moment propagation for diffusion models, placing a last-layer Laplace posterior~\eqref{eq:llla} on the denoiser and carrying the mean and variance of the state along the sampling chain, so that the terminal state variance captures the epistemic uncertainty of the generated sample. Here we express its recursion for the flow matching sampler~\eqref{eq:euler}, where different weights drawn from the posterior can predict different velocities at the same state $x$ and time $t$. BayesDiff represents uncertainty through per-coordinate variances, retaining only the diagonal of the predictive covariance:
\begin{equation}
\label{eq:predictive}
p\!\left(u \mid x, t, \D\right) \approx \Normal\!\left(u;\, u_{\thetaMAP}(x, t),\, \diag \gamma^{2}(x, t)\right),
\end{equation}
\noindent
with mean velocity produced by the MAP weights and variance $\gamma^{2}(x, t)$ given by the diagonal of $J^{\theta} \Sigma_\theta (J^{\theta})^\transpose$, where $J^{\theta}(x, t) \triangleq \partial u / \partial \theta$ is the Jacobian of the velocity with respect to the weights. Writing $u_i \triangleq u_{\theta}(x_i, t_i)$ for the velocity at step $i$, the procedure carries the mean $m_i \triangleq \E[x_i]$ and the covariance of the state $P_i \triangleq \Cov(x_i)$, where all expectations, variances and covariances are taken over $x_i$ and $u_i$. Taking expectations and variances of the Euler update~\eqref{eq:euler}, which is affine in the velocity, gives an exact moment recursion. BayesDiff keeps the mean and diagonal of $P_i$:
\begin{align}
m_{i+1} &= m_i + \Delta t\, \E\!\left[u_i\right],
\label{eq:mean_recursion}\\
P_{i+1} &= P_i
  + 2\,\Delta t\, \diag \Cov\!\left(x_i, u_i\right)
  + \Delta t^{2}\, \diag \Cov\!\left(u_i\right),
\label{eq:var_recursion}
\end{align}
\noindent
started from $m_0 = x_0$ and $P_0 = 0$ and iterated for $T$ steps, with final state covariance $P_T$ estimating the uncertainty of the generated sample. BayesDiff takes the conditional velocity variance from~\eqref{eq:predictive} and estimates the cross-covariance $\Cov(x_i, u_i)$ by Monte Carlo sampling $S$ particles from $x_i^{(s)} \sim \Normal(m_i, P_i)$ and evaluating their MAP velocities. Every expectation above is taken under the product $p(u \mid x_i, t_i, \D)\, p(x_i)$, whose first factor is the predictive distribution~\eqref{eq:predictive} with the weights integrated out, $p(u \mid x_i, t_i, \D) = \int p(u \mid x_i, t_i, \theta)\, p(\theta \mid \D)\, \dd\theta$. However, this marginalization treats the weights as independent of the state they generate, which is the key limitation of BayesDiff. A posterior trajectory uses one weight draw throughout integration, so $x_i$ depends on that draw. Discarding this dependence leads to the variance collapse shown in the next section.

\section{Coherent Moment Propagation}
\label{sec:comp}

In this section, we prove that BayesDiff's estimated total variance collapses to zero as the integration grid of the solver is refined. We then introduce Coherent Moment Propagation (CoMP), which avoids this collapse by propagating the state-weight cross-covariance through the ODE sampler.

\paragraph{BayesDiff total variance recursion.}  To analyze BayesDiff's reported uncertainty, we consider the \emph{total variance} of the generated sample, $\Tr P_T = \sum_d (P_T)_{dd}$. Taking the trace of~\eqref{eq:var_recursion} collapses the covariance update into a single scalar recursion,
\begin{equation}
\label{eq:identity}
\Tr P_{i+1}
= \underbrace{\Tr \Cov_x\!\left[x_i + \Delta t\, u_{\thetaMAP}(x_i, t_i)\right]}_{\text{transport of the belief}}
+ \underbrace{\Delta t^{2}\, \bar\sigma_i^{2}}_{\text{epistemic injection}},
\end{equation}
\noindent
where $\E_x \equiv \E_{x_i}$, $\Cov_x \equiv \Cov_{x_i}$, and $\bar\sigma_i^{2} \triangleq \E_x[\sum_d \gamma_d^{2}(x_i,t_i)]$ averages the summed predictive variance~\eqref{eq:predictive} over the state belief. Each step pushes the belief through the deterministic Euler map $h_i(x_i) \triangleq x_i + \Delta t\, u_{\thetaMAP}(x_i, t_i)$ and adds the epistemic injection $\Delta t^{2} \bar\sigma_i^{2}$, through which the weight posterior enters the update. Since the chain takes $T = 1/\Delta t$ steps, the epistemic content of the reported spread is $T$ injections of size $\Delta t^{2}$, and whether anything survives the refinement of the integration grid depends entirely on how much the transport term amplifies each contribution.

\paragraph{The estimated total variance is upper bounded by $O(1/T)$.}
We prove that BayesDiff's total variance is upper bounded by $O(1/T)$, so increasing the number of integration steps leads to an underestimated uncertainty. The result uses the joint law of~\eqref{eq:var_recursion}, a fixed initial noise sample ($\Tr P_0 = 0$), and two regularity conditions:\looseness-1
\begin{itemize}[nosep, topsep=4pt, leftmargin=3.2em, labelwidth=2.6em, labelsep=0.6em, align=left]
  \item[\textbf{(A1)}] \emph{Lipschitz velocity field.} The velocity field $u_{\thetaMAP}$ is globally Lipschitz in the state with constant $L$, as is normally assumed when analyzing the convergence of diffusion and flow samplers~\citep{chen2023probability, benton2024error}.
  \item[\textbf{(A2)}] \emph{Bounded noise injection rate.} The summed predictive variance obeys $\sum_{d} \gamma^{2}_d(x, t) \le \bar\sigma_{\max}^{2} < \infty$ at every state and time, with a bound independent of $T$ such that $\bar\sigma_i^{2} \le \bar\sigma_{\max}^{2}$.
\end{itemize}

\begin{theorem}[Variance collapse]
\label{thm:decay}
Under (A1) and (A2), the total variance under moment recursion~\eqref{eq:var_recursion} satisfies
\begin{equation}
\label{eq:decay}
\Tr P_T \;\le\; e^{2L}\, \frac{\bar\sigma_{\max}^{2}}{T} \;=\; O(1/T),
\end{equation}
\noindent
i.e., the total variance vanishes as the integration grid refines. See Appendix~\ref{app:decay} for the proof.
\end{theorem}

\noindent
\emph{Proof sketch.} Each step of~\eqref{eq:identity} stretches the belief by at most $(1 + \Delta t\, L)^{2}$ under (A1) and adds at most $\Delta t^{2} \bar\sigma_{\max}^{2}$ under (A2), so the $T$ injections are bounded by $e^{2L} \bar\sigma_{\max}^{2} / T$. Importantly, under (A1) and Lipschitz continuity in time, the Euler iterates converge to the solution of the continuous-time ODE $\dd x / \dd t = u_{\thetaMAP}(x, t)$ as $T \to \infty$ at first order in the step size~\citep{hairer1993solving}. Corollary~\ref{cor:fullcov} shows that carrying full state covariance does not prevent the collapse. We also extend this bound to diffusion SDEs: Corollary~\ref{cor:sde} gives an $O(1/T)$ epistemic term in the upper bound alongside an $O(1)$ aleatoric term from Brownian noise (Appendix~\ref{app:decay}). A finer integration grid can therefore collapse the reported total variance, underestimating epistemic uncertainty. To overcome this limitation, we introduce an alternative to BayesDiff called CoMP, which restores the omitted state-weight cross-covariance, giving an upper bound on total variance independent of the number of integration steps.

Our key idea is that rather than marginalize the weights at every step, we linearize the ODE sampler~\eqref{eq:euler} about its mean trajectory and propagate the state and the weight deviation jointly, so the recursion accounts for their dependence instead of implicitly setting it to zero. We linearize jointly in $(x_i,\theta)$ about $(m_i,\thetaMAP)$, where (A1) bounds the state Jacobian by $\|J^{x}_i\| \le L$. For a linear last layer\footnote{The action heads of recent VLAs are of this form, among them $\pi_{0.5}$~\citep{intelligence2025pi05}, GR00T~\citep{bjorck2025gr00t} and SmolVLA~\citep{shukor2025smolvla}, each of which has a linear last layer.} the parameter Jacobian is available analytically. We develop two approximations for propagating state uncertainty through the ODE sampler. 

Expanding the Euler update $x_i + \Delta t\, u_\theta(x_i,t_i)$ to first order in both $x_i$ and $\theta$ about $(m_i,\thetaMAP)$ gives the linearized recursion
\begin{equation}
\label{eq:comp_lin}
\delta x_{i+1} = A_i\, \delta x_i + B_i\, \delta\theta,
\qquad
A_i \triangleq I + \Delta t\, J^{x}_i,
\qquad
B_i \triangleq \Delta t\, J^{\theta}_i,
\end{equation}
\noindent
with $\delta x_i \triangleq x_i - m_i$ the state deviation under the linearized model, $\delta\theta$ the weight deviation of \S\ref{sec:setup}, drawn once and held fixed throughout integration, and $J^{x}_i$ and $J^{\theta}_i$ the state and weight Jacobians of the velocity at $(m_i, t_i, \thetaMAP)$. Stacking the two deviations into $z_i \triangleq (\delta x_i, \delta\theta)$ turns~\eqref{eq:comp_lin} into the linear map $z_{i+1} = F_i z_i$ with $F_i$ the block matrix with rows $(A_i, B_i)$ and $(0, I)$, whose identity block carries $\delta\theta$ unchanged, so the joint moments propagate in closed form. We carry the state mean $m_i$, the full $D \times D$ covariance $P_i$, and the state-parameter cross-covariance $
C_i \triangleq \Cov(\delta x_i,\, \delta\theta)$,
which measures how much of the state's departure by step $i$ is explained by the weights that drove it there. For this linearized model, the moment updates are exact. Written blockwise, Coherent Moment Propagation carries the triple $(m_i, P_i, C_i)$ from $(x_0, 0, 0)$ to the end of the chain by
\begin{align}
m_{i+1} &= h_i(m_i),
\label{eq:comp_mean}\\
C_{i+1} &= A_i\, C_i + B_i\, \Sigma_\theta,
\label{eq:comp_cross}\\
P_{i+1} &= \underbrace{A_i P_i A_i^{\transpose}}_{\text{transported state spread}}
+ \underbrace{A_i C_i B_i^{\transpose} + B_i C_i^{\transpose} A_i^{\transpose}}_{\text{state-weight cross terms}}
+ \underbrace{B_i \Sigma_\theta B_i^{\transpose}}_{\text{weight contribution}} .
\label{eq:comp_state}
\end{align}
\noindent
The mean~\eqref{eq:comp_mean} follows the MAP trajectory;~\eqref{eq:comp_cross} supplies the cross terms. At a fixed state, a linear last layer makes the velocity affine in $\theta$, so $J^{\theta}_i$ and $B_i$ can be computed analytically from its features.

\paragraph{The upper bound does not depend on the number of integration steps.}
The same two assumptions that gave Theorem~\ref{thm:decay} its upper bound give this recursion a different one, under the same integration grid and the same constant.

\begin{theorem}[Upper bound independent of integration steps]
\label{thm:stable}
Under (A1) and (A2), the total variance under recursion~\eqref{eq:comp_mean}--\eqref{eq:comp_state} from $(x_0, 0, 0)$ satisfies
\begin{equation}
\label{eq:stable}
\Tr P_T \;\le\; e^{2L}\, \bar\sigma_{\max}^{2} \;=\; O(1),
\end{equation}
\noindent
whereas the same recursion with $C_i\equiv0$ has the same upper bound~\eqref{eq:decay} as Theorem~\ref{thm:decay}.
See Appendix~\ref{app:coherence} for the proof.
\end{theorem}
\noindent
The two upper bounds carry the same constant and differ by the factor $\Delta t = 1/T$. Unrolling the recursion collects $T$ contributions of size $\Delta t$: carrying $C_i$ adds them as amplitudes before squaring, while holding $C_i$ at zero adds their squares as variances. We therefore attribute the vanishing spread of Theorem~\ref{thm:decay} to the cross-covariance that BayesDiff omits, and Theorem~\ref{thm:stable} says that carrying it removes the factor $1/T$ from the bound. Figure~\ref{fig:shrinkage} compares the estimators' spread with that of a posterior ensemble on a flow matching model of a two-dimensional mixture. To realize this estimator, we still need the state transition $A_i$ and transported spread $A_i P_i A_i^{\transpose}$. We derive two variants inspired by the extended and unscented Kalman filters.

\paragraph{CoMP-EKF.} With the state Jacobian $J^{x}_i = \partial u_{\thetaMAP} / \partial x_i$ evaluated at $(m_i, t_i)$, running~\eqref{eq:comp_mean}--\eqref{eq:comp_state} as written is the extended Kalman filter's predict step~\citep{sarkka2013bayesian} applied to the sampler, with batched autodiff supplying $J^{x}_i$ at each step. We provide pseudocode in Algorithm~\ref{alg:ekf}.

\paragraph{CoMP-UKF.}
The unscented transform~\citep{sarkka2013bayesian} represents the state mean $m_i$ and covariance $P_i$ with $2D+1$ deterministic sigma points, which we propagate through the nonlinear Euler map $h_i$, capturing nonlinear effects beyond CoMP-EKF's local linearization. The same sigma points estimate $A_i$ for the state-weight cross-covariance update in~\eqref{eq:comp_cross}. Appendix~\ref{app:algo_ekf} gives the moment equations and pseudocode (Algorithm~\ref{alg:ukf}). Both CoMP estimators use the trace of $P_T$ or of the chunk's endpoint covariance as the \textbf{uncertainty score} $U$ (Appendix~\ref{app:score}).

\paragraph{Augmented head.}
We run CoMP with the plain last-layer posterior over features $\phi$ and the augmented posterior~\eqref{eq:cond_head}. LULA~\citep{kristiadi2021lula} learns the added features using an out-of-distribution set, which we do not assume is available. We instead set $\zeta = c(o)$ in~\eqref{eq:cond_head}, reusing the image, language and proprioception embeddings already computed by VLAs such as $\pi_{0.5}$~\citep{intelligence2025pi05} and SmolVLA~\citep{shukor2025smolvla}. Fitting this augmented Laplace posterior requires no policy retraining and it uses $\psi$ instead of $\phi$ for $B_i$ in~\eqref{eq:comp_cross}--\eqref{eq:comp_state}.

\section{Evaluation}
\label{sec:evaluation}

To assess how CoMP's uncertainty estimates benefit pre-trained policies, we ask four questions:

\begin{itemize}[nosep, topsep=3pt, leftmargin=2.8em, labelwidth=2.2em, labelsep=0.6em, align=left]
  \item[\textbf{Q1:}] Does the covariance our estimator propagates match the posterior it approximates? (\S\ref{sec:calibration})
  \item[\textbf{Q2:}] Does the augmented head make uncertainty estimates better reflect input ambiguity, particularly in language instructions? (\S\ref{sec:conditioning})
  \item[\textbf{Q3:}] Can CoMP's uncertainty estimates predict future failures during policy execution? (\S\ref{sec:failure})
  \item[\textbf{Q4:}] Can uncertainty-guided demonstration selection improve sample efficiency for policy fine-tuning in new tasks? (\S\ref{sec:finetuning})
\end{itemize}

\paragraph{Models and baselines.}
We evaluate three robot policies: a transformer-based Diffusion Policy~\citep{chi2023diffusion} trained with flow matching~\citep{lipman2023flow} on robomimic~\citep{mandlekar2021robomimic}, $\pi_{0.5}$~\citep{intelligence2025pi05} on LIBERO~\citep{liu2023libero} and RoboLab~\citep{yang2026robolab}, and SmolVLA~\citep{shukor2025smolvla} on LIBERO. For $\pi_{0.5}$ and SmolVLA, we retain the default number of integration steps except in ablations (Appendix~\ref{app:implementation}). We adapt five post-hoc uncertainty estimators to flow matching policies. \textbf{BayesDiff}~\citep{kou2024bayesdiff} is the Monte Carlo moment recursion described in \S\ref{sec:setup}. \textbf{DEUA}~\citep{huang2026deua} re-noises the generated sample and scores velocity variance across posterior draws. \textbf{GU}~\citep{jazbec2025generative} is the Gaussian entropy of the terminal state across posterior draws. \textbf{DECU}~\citep{berry2024shedding} estimates mutual information from the pairwise distances between member velocities at a branching point of the integration path. \textbf{VFD}~\citep{romer2026save} integrates the time-weighted disagreement between member velocity fields along the integration path. We also report DECU and VFD with independently trained ensembles as references~\citep{romer2026save}, since they require policy training beyond our test-time setting. All post-hoc estimators use Kronecker-factored last-layer Laplace posteriors~\eqref{eq:llla}. 

\subsection{Faithfulness of the propagated covariance}
\label{sec:calibration}

To answer Q1, we test the calibration of the moment propagation estimators against a reference chunk covariance, $P^{\star}$, estimated via posterior ensembles on robomimic~\citep{mandlekar2021robomimic}. We train one transformer-based policy per task on the proficient-human demonstrations of \textit{Lift}, \textit{PickPlaceCan} and \textit{NutAssemblySquare} (Figure~\ref{fig:robomimic_perturb}). The posterior~\eqref{eq:llla} is fit on the training split; training and sampler settings are given in Appendix~\ref{app:faithfulness}. We use three complementary metrics to assess calibration. The \textbf{total variance ratio} $\Tr P / \Tr P^{\star}$ measures recovered spread. The \textbf{variogram discrepancy} compares pairwise moments of order $1/2$~\citep{scheuerer2015variogram} and is sensitive to correlations. Both metrics are reported relative to their reference values, so a value of $1$ corresponds to a perfect match. Total variance ratios below or above $1$ indicate under- or overestimation, respectively, while variogram ratios above $1$ indicate greater discrepancy with held-out posterior samples than the reference Gaussian. Finally, we use the \textbf{negative log likelihood (NLL)} to measure how much density the estimated Gaussian assigns to expert demonstrations; lower is better.

\begin{figure}[t]
  \centering
  \includegraphics[width=\textwidth]{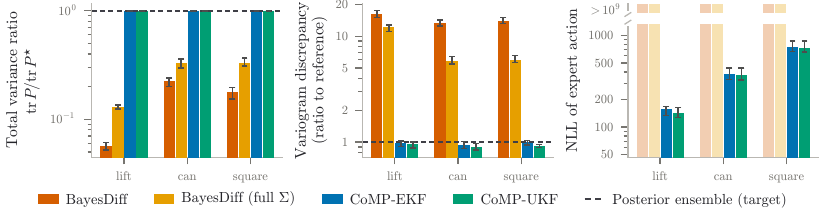}
  \caption{\textbf{Calibration against the posterior ensemble on robomimic.} Each bar is
  the median over $150$ observation windows per task with a bootstrap $95\%$ interval. Reference values in the first two panels use $M = 6000$ posterior draws. CoMP outperforms BayesDiff across all three metrics, with expert-action negative log likelihoods (NLLs) far below those of both BayesDiff variants ($>10^9$). \rebuttal{In Figure~\ref{fig:ranking_study} we show that the number of Euler steps not only affects the scale of BayesDiff's total variance but also changes its ranking of the most and least uncertain observations. BayesDiff ranks at most $40\%$ of these observations correctly and reorders them at every evaluated step count, whereas CoMP ranks $77$ to $93\%$ correctly with a stable ordering across different numbers of integration steps.}
  } \label{fig:calibration}
\end{figure}

Figure~\ref{fig:calibration} shows results with the plain head\footnote{This keeps the focus on covariance propagation; augmented-head results are similar (Figure~\ref{fig:calibration_aug}).}. BayesDiff recovers only $6\%$ to $23\%$ of the reference variance depending on the task (Figure~\ref{fig:calibration}, left), and its variogram discrepancy is $13$ to $16$ times the reference value (Figure~\ref{fig:calibration}, middle), i.e., the correlation across the chunk is misreported along with its scale.  Propagating the full covariance with BayesDiff, i.e., \textit{BayesDiff (full $\Sigma$)}, improves over \textit{BayesDiff} but still misestimates the posterior distribution, as shown in Corollary~\ref{cor:fullcov}. Both CoMP estimators recover the reference variance to within $0.2\%$ and variogram discrepancy to within $10\%$, with lower NLLs indicating better agreement with expert demonstrations, showing better calibration overall. \looseness-1 

\subsection{Response of the augmented head to the observation}
\label{sec:conditioning}

\begin{wrapfigure}{R}{0.4\textwidth}
  \vspace{-1.1\baselineskip}
  \centering
  \includegraphics[width=0.4\textwidth]{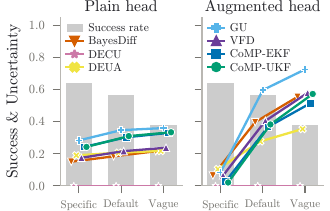}
  \caption{\textbf{Language instruction shift on RoboLab.} Success rate (grey bars) and first-decision uncertainty (lines, scaled to $[0,1]$) averaged over the ten tasks by wording. On the plain head (left) estimators are mostly flat with the wording; on the augmented head (right) every estimator but DECU rises as language ambiguity increases.}
  \label{fig:robolab_shift}
  \vspace{-1\baselineskip}
\end{wrapfigure}

To test whether the augmented head improves the response of estimated uncertainty to policy inputs, e.g., language instructions (Q2), we use RoboLab~\citep{yang2026robolab} for its high visual fidelity and three levels of instruction specificity: specific, default and vague (see Table~\ref{tab:robolab_wordings}). We evaluate ten tabletop tasks (Figure~\ref{fig:robolab_scenes}) with $100$ rollouts per task and wording. We use openpi's $\pi_{0.5}$ fine-tuned on DROID~\citep{khazatsky2024droid}, as released, which has \emph{not} been trained on RoboLab data. We fit both Laplace heads, plain head and augmented head, from \S\ref{sec:comp}, once for all ten tasks, on up to ten successful rollouts of the \emph{specific} wording mixed with DROID data. We evaluate uncertainty using only the first observation of each held-out rollout, varying only the instruction wording while keeping images and proprioception fixed. See Appendix~\ref{app:robolab_tasks} for training details.\looseness-1

As instructions become less specific, success rates should decline and uncertainty increase. However, all estimators report near-constant uncertainty with the plain head (see Figure~\ref{fig:robolab_shift}). With the augmented head, uncertainty rises with instruction ambiguity, for every estimator but DECU, which is saturated at $\ln M$, the ceiling of its mutual-information proxy (per-task results in Figure~\ref{fig:robolab_shift_tasks}). This response shows that the augmented head identifies ambiguous language instructions with the vision-language embeddings readily computed by the pre-trained policy. We note that BayesDiff's responsiveness to ambiguous instructions decreases with more integration steps, whereas CoMP is not affected (Figure~\ref{fig:robolab_shift_steps}), validating our theoretical results in \S\ref{sec:comp}.

\subsection{Runtime failure detection}
\label{sec:failure}

To evaluate early failure detection (Q3), we follow FIPER~\citep{romer2025fiper, romer2026save} on two simulation suites. On robomimic, we use the same tasks and policies of \S\ref{sec:calibration} with perturbed initial end-effector and object poses (Figure~\ref{fig:robomimic_perturb}), with $50$ rollouts per task and perturbation condition. On LIBERO~\citep{liu2023libero}, we evaluate openpi's \texttt{pi05\_libero} on nine held-out LIBERO-90 tasks (Figure~\ref{fig:libero90_scenes}), chosen with a success rate between $0.3$ and $0.7$ as in FIPER.  Motivated by the results from \S\ref{sec:conditioning}, we use CoMP with the augmented head and compare it against post-hoc baselines\footnote{The post-hoc baselines use the plain head by default, though we also evaluate them with our augmented head in Appendix~\ref{app:fiper_tables} to isolate its contribution.}.\looseness-1 

\begin{figure}[t]
  \centering
  \includegraphics[width=\textwidth]{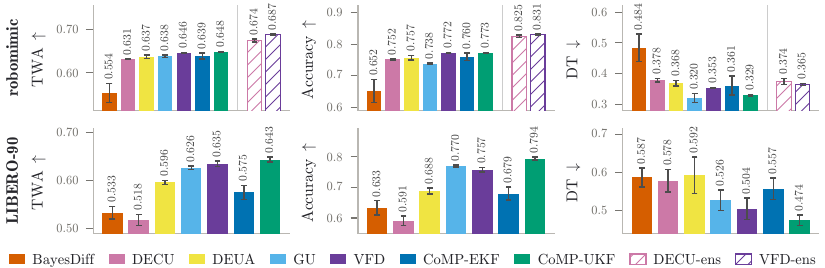}
  \caption{\textbf{Runtime failure detection on robomimic and LIBERO-90.} Timestep-wise accuracy (TWA), accuracy and normalized detection time (DT), averaged over tasks, with mean and standard deviation over five random seeds.  CoMP uses our augmented head; post-hoc baselines use the plain head. The ensembles (hatched, two trained members) are a reference only. \rebuttal{CoMP-UKF is comparable to or better than the baselines in both suites}; Table~\ref{tab:fiper_heads} reports comparisons with both heads.\looseness-1} \label{fig:fiper_main}
\end{figure}

Following~\cite{romer2025fiper}, a rollout is flagged once the uncertainty score $U$ exceeds a conformal threshold~\citep{angelopoulos2023conformal} calibrated on $20$ rollouts per task. We report detection accuracy, detection time normalized by the maximum episode length (DT), and timestep-wise accuracy (TWA), an accuracy measure that penalizes late failure detections. \rebuttal{CoMP-UKF matches the best post-hoc baseline in TWA and accuracy on robomimic. On LIBERO-90, it achieves the highest TWA ($0.643$), accuracy ($0.794$) and lowest DT ($0.474$),} while CoMP-EKF lags behind (Figure~\ref{fig:fiper_main}), which we attribute to linearization errors. \rebuttal{The same holds when all baselines use the augmented head (Table~\ref{tab:fiper_heads}): CoMP-UKF keeps the best TWA on robomimic, is within $0.002$ of DECU in accuracy, and leads all three metrics on LIBERO-90. Both CoMP variants improve over BayesDiff on every metric with either head, and plain-head CoMP-UKF matches the best plain-head baseline.}
Although the trained ensembles (hatched) lead on robomimic, CoMP remains competitive at a fraction of their fitting and storage cost: its Laplace posterior stores $\lesssim 1\%$ of the ensemble parameters, fit in one pass over the training split (see Appendix~\ref{app:fiper_tables} for implementation details and additional results).\looseness-1

\subsection{Uncertainty-guided fine-tuning}
\label{sec:finetuning}

\begin{wrapfigure}{R}{0.4\textwidth}
  \vspace{-1.1\baselineskip}
  \centering
  \includegraphics[width=0.4\textwidth]{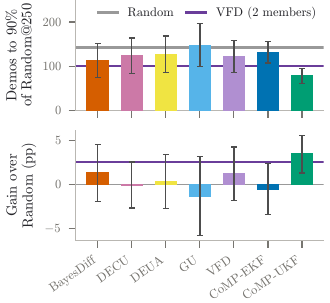}
  \caption{\textbf{Data selection on LIBERO-10.} Demonstrations needed to reach $90$\% of random's success at $X = 250$ (top) and mean gain over random (bottom). CoMP uses the augmented head; post-hoc baselines use the plain head.
  }
  \label{fig:oneshot_gain}
  \vspace{-0.8\baselineskip}
\end{wrapfigure}

To test whether a pre-trained policy's uncertainty can inform demonstration selection for efficient fine-tuning (Q4), we adapt the budgeted selection setting of CUPID~\citep{agia2025cupid} and FAKTUAL~\citep{sirigiri2026diversity}: we compare policies fine-tuned on $X$ demonstrations selected by uncertainty or drawn at random. Following SAVE~\citep{romer2026save}, we train SmolVLA~\citep{shukor2025smolvla} on the LIBERO-Spatial, Object and Goal suites~\citep{liu2023libero} and three LIBERO-10 tasks; all LIBERO-10 demonstrations form the pool (Figure~\ref{fig:libero10_scenes}). Each estimator scores demonstrations at their first frame for uncertainty-guided sampling. We fine-tune once on the selection mixed with the original training data and evaluate on all ten tasks. We repeat this for five independently trained policies and budgets $X$ from $25$ to $250$, with a two-member VFD as a reference. We compare CoMP with the same set of post-hoc baselines as in \S\ref{sec:failure}. Appendix~\ref{app:oneshot} presents results of post-hoc baselines with both heads (see Figure~\ref{fig:oneshot_heads}) and provides implementation details.

Figure~\ref{fig:oneshot_gain} shows how many demonstrations each method needs to reach $90\%$ of the success achieved with $X = 250$ randomly selected demonstrations (top), and its mean success-rate gain over random selection across budgets (bottom). On average, CoMP-UKF reaches the target success rate with \rebuttal{fewer demonstrations than the plain-head baselines} and yields \rebuttal{a larger mean gain over random selection ($3.5$ pp) than the two-member VFD ensemble}, using a single policy and a $75$ MB posterior instead of a second $1.2$ GB policy. CoMP-EKF shows little benefit, possibly missing nonlinear effects that CoMP-UKF captures by propagating sigma points through the network. In the supplementary material we show how our augmented head improves fine-tuning results for every estimator (Figure~\ref{fig:oneshot_heads}), addressing the limited expressivity of the plain head\rebuttal{, with several baselines reaching gains within one standard deviation of CoMP-UKF's. We also show that BayesDiff loses its fine-tuning gain as integration steps increase: with the augmented head, its gain falls from $2.7$ to $-2.9$~pp between $10$ and $50$ steps, while CoMP-UKF's holds at $3.3$ to $3.7$~pp (Table~\ref{tab:oneshot_steps}).} Metric definitions and full curves are in Appendix~\ref{app:oneshot}.

\section{Related Work}
\label{sec:related}

Epistemic uncertainty helps neural networks quantify when their own predictions may be unreliable~\citep{kendall2017uncertainties} and is commonly estimated with Bayesian weight approximations~\citep{blundell2015weight, gal2016dropout, maddox2019simple} or ensembles~\citep{lakshminarayanan2017deep, pearce2020uncertainty}. In diffusion and flow matching models, ensembling is a common approach to estimating uncertainty~\citep{berry2024shedding, romer2026save}, while other methods use Bayesian hypernetworks~\citep{krueger2017hypernetworks, chan2024estimating}, disagreement between existing expert paths~\citep{berry2025emoe}, or Laplace posteriors~\citep{jazbec2025generative, huang2026deua}. Other uncertainty proxies include input sensitivity, velocity-field divergence and velocity variance~\citep{de2024diffusion, xing2026divergence, de2026flowing}. We focus on uncertainty estimation from a single pre-trained flow policy. 

Moment propagation instead carries moments through denoising. BayesDiff~\citep{kou2024bayesdiff} applied it to diffusion models with a last-layer Laplace posterior~\citep{kristiadi2020llla, daxberger2021laplace}, followed by extensions to trajectory diffusion~\citep{capellera2025unified} and flow matching~\citep{han2026flow}. However, these approximations have limited expressivity and are susceptible to feature collapse~\citep{vanamersfoort2021feature, mukhoti2023ddu}. Extensions use broader parameter coverage~\citep{gupta2026quantifying} or neural tangent kernel approximations~\citep{deng2022ella, calvo2026richer}, while LULA~\citep{kristiadi2021lula} learns additional units without changing the model's predictions. CoMP addresses BayesDiff's variance shrinkage by retaining the state-weight cross-covariance, while our LULA-inspired widening reuses observation embeddings without policy re-training.

Uncertainty quantification helps robots anticipate failures and decide what to learn next. Runtime monitors use uncertainty or action consistency to flag failures early~\citep{romer2025fiper, xu2025faildetect, agia2024sentinel}. To adapt policies with fewer demonstrations, uncertainty guides selection through information gain~\citep{bagatella2025active} or velocity-field disagreement (VFD)~\citep{romer2026save}. Complementary criteria include influence functions~\citep{agia2025cupid}, datamodels~\citep{dass2026datamil} and trajectory diversity~\citep{sirigiri2026diversity}. CoMP supports these applications with uncertainty extracted from pre-trained foundation policies without re-training them.

\section{\rebuttal{Limitations}}
\label{sec:limitations}

\begin{table}[t]
  \centering
  \footnotesize
  \caption{\rebuttal{\textbf{Inference cost at the settings of \S\ref{sec:failure} and \S\ref{sec:finetuning}.} Wall time per decision and peak GPU memory on one A100 GPU for $\pi_{0.5}$ on LIBERO-90 ($D = 70$, $90$ test episodes) and SmolVLA on LIBERO-10 ($D = 350$, $379$ demonstrations).\looseness-1}} 
  \label{tab:cost}
  {\color{rebuttalblue}\begin{tabular}{@{}llccccccc@{}}
\toprule
 & & \multicolumn{5}{c}{Post-hoc estimators} & \multicolumn{2}{c}{Ours} \\
\cmidrule(lr){3-7} \cmidrule(lr){8-9}
 & & BayesDiff & DEUA & GU & DECU & VFD & CoMP-EKF & CoMP-UKF \\
\midrule
\multirow{2}{*}{Time (s)} & $\pi_{0.5}$ & 0.91 & 0.50 & 0.94 & 0.53 & 1.80 & 2.13 & 1.85 \\
 & SmolVLA & 0.98 & 0.54 & 0.99 & 0.59 & 1.11 & 10.98 & 3.38 \\
\midrule
\multirow{2}{*}{Memory (GB)} & $\pi_{0.5}$ & 7.8 & 7.7 & 7.9 & 7.7 & 10.5 & 8.3 & 10.8 \\
 & SmolVLA & 1.4 & 1.3 & 1.3 & 1.3 & 1.9 & 1.9 & 4.6 \\
\bottomrule
\end{tabular}}
  \vspace{-0.5\baselineskip}
\end{table}

Despite its benefits, CoMP suffers from \rebuttal{three} limitations: it relies on a linear last layer for analytic weight Jacobians, making deeper Laplace posteriors costly, and its state covariance storage scales quadratically with the output dimension. \rebuttal{The third is its inference cost, which grows with the flattened chunk dimension $D = H d_a$: CoMP-UKF evaluates $2D + 1$ sigma points per Euler step and CoMP-EKF needs $D$ backward passes to compute the state Jacobian. In Table~\ref{tab:cost}, CoMP-UKF takes $1.85$~s per decision and $10.8$~GB of peak memory on $\pi_{0.5}$, close to VFD ($1.80$~s and $10.5$~GB), one of the strongest baselines. However, on SmolVLA, whose chunk dimension is five times that of $\pi_{0.5}$ (due to $H = 50$ predicted actions instead of $10$), CoMP-UKF takes $3.38$~s and $4.6$~GB, three times the time and more than twice the memory of VFD, since the sigma points are evaluated as one batch through the action expert. As expected, CoMP-EKF is the most expensive estimator on both policies due to its Jacobian computations, although its memory stays within the range of the baselines.\looseness-1} 

\rebuttal{Nevertheless, in offline settings such as demonstration selection (\S\ref{sec:finetuning}), and in safety-critical deployments where a missed failure costs more than a slower decision, this additional computation is justified. Moreover, the gains of our variance propagation are not an artifact of the number of network evaluations: a Laplace posterior ensemble with the same number of draws as CoMP-UKF has sigma points is still worse on LIBERO-90 (Table~\ref{tab:sampling_budget}), highlighting the value of our estimator. Note that slow inference is not specific to CoMP: every post-hoc estimator in Table~\ref{tab:cost} needs at least $0.5$~s per decision (DEUA), leaving real-time uncertainty estimation for robot policies as an open problem.} These limitations motivate low-rank covariance propagation and scalable posterior approximations beyond the last layer, which could broaden CoMP's applications, including to world action models and long-horizon assembly tasks.\looseness-1

\section{Conclusion}
\label{sec:conclusion}

We present Coherent Moment Propagation (CoMP), an epistemic uncertainty estimator for pre-trained flow matching-based robot policies. We develop two novel uncertainty estimators, CoMP-EKF and CoMP-UKF, that propagate the state-weight cross-covariance through the ODE sampler. Through theoretical and empirical analysis, we show that BayesDiff underestimates uncertainty by omitting this cross-covariance. Specifically, its reported total variance is upper bounded by $O(1/T)$ in the number of Euler steps $T$, a phenomenon we call variance collapse: refining the solver drives uncertainty estimates to zero even as numerical accuracy of the solver improves. Inspired by LULA~\citep{kristiadi2021lula}, we further propose to augment the last-layer features with vision-language embeddings readily computed by pre-trained policies, and show that the augmented head improves uncertainty estimators' responsiveness to policy input without re-training the policy. \rebuttal{Experiments} across three simulated robotics benchmarks show that CoMP avoids BayesDiff's variance collapse and is on par with or better than established post-hoc uncertainty estimators on applications including failure detection and efficient fine-tuning.\looseness-1

\subsection*{AI use statement}

We used generative AI tools to polish the manuscript, write unit tests for our methods and baselines, create experiment launch scripts, check the consistency of mathematical notation, and improve figure presentation. We reviewed the AI-assisted material and take responsibility for the final manuscript.

\subsection*{Reproducibility statement}

We provide proofs of our theoretical results in Appendices~\ref{app:decay} and~\ref{app:coherence}, pseudocode in Appendix~\ref{app:algo_ekf}, and experimental settings and implementation details in Appendices~\ref{app:implementation}--\ref{app:oneshot}. The code used to produce our results will be released publicly following publication.

\newpage

\bibliographystyle{iclr2027_conference}
\bibliography{references}

@inproceedings{daxberger2021laplace,
	title        = {Laplace Redux - Effortless Bayesian Deep Learning},
	author       = {Erik Daxberger and Agustinus Kristiadi and Alexander Immer and Runa Eschenhagen and Matthias Bauer and Philipp Hennig},
	year         = 2021,
	booktitle    = {Advances in Neural Information Processing Systems},
	editor       = {A. Beygelzimer and Y. Dauphin and P. Liang and J. Wortman Vaughan}
}

@article{ho2020ddpm,
	title        = {Denoising diffusion probabilistic models},
	author       = {Ho, Jonathan and Jain, Ajay and Abbeel, Pieter},
	year         = 2020,
	journal      = {Advances in neural information processing systems},
	volume       = 33,
	pages        = {6840--6851}
}

@inproceedings{huang2026deua,
	title        = {Diffusion epistemic uncertainty with asymmetric learning for diffusion-generated image detection},
	author       = {Huang, Yingsong and Guo, Hui and Huang, Jing and Bai, Bing and Xiong, Qi},
	year         = 2025,
	booktitle    = {2025 IEEE/CVF International Conference on Computer Vision (ICCV)},
	pages        = {17097--17107},
	organization = {IEEE}
}

@inproceedings{kou2024bayesdiff,
	title        = {{BayesDiff}: Estimating pixel-wise uncertainty in diffusion via bayesian inference},
	author       = {Kou, Siqi and Gan, Lei and Wang, Dequan and Li, Chongxuan and Deng, Zhijie},
	year         = 2024,
	booktitle    = {International Conference on Learning Representations},
	volume       = 2024,
	pages        = {17046--17063}
}

@inproceedings{capellera2025unified,
	title        = {Unified uncertainty-aware diffusion for multi-agent trajectory modeling},
	author       = {Capellera, Guillem and Rubio, Antonio and Ferraz, Luis and Agudo, Antonio},
	year         = 2025,
	booktitle    = {2025 IEEE/CVF Conference on Computer Vision and Pattern Recognition (CVPR)},
	pages        = {22476--22486},
	organization = {IEEE}
}

@inproceedings{kristiadi2020llla,
	title        = {Being bayesian, even just a bit, fixes overconfidence in {ReLU} networks},
	author       = {Kristiadi, Agustinus and Hein, Matthias and Hennig, Philipp},
	year         = 2020,
	booktitle    = {International conference on machine learning},
	pages        = {5436--5446},
	organization = {PMLR}
}

@inproceedings{lipman2023flow,
	title        = {Flow Matching for Generative Modeling},
	author       = {Yaron Lipman and Ricky T. Q. Chen and Heli Ben-Hamu and Maximilian Nickel and Matthew Le},
	year         = 2023,
	booktitle    = {The Eleventh International Conference on Learning Representations}
}

@inproceedings{liu2023flow,
	title        = {Flow Straight and Fast: Learning to Generate and Transfer Data with Rectified Flow},
	author       = {Xingchao Liu and Chengyue Gong and qiang liu},
	year         = 2023,
	booktitle    = {The Eleventh International Conference on Learning Representations}
}

@inproceedings{song2021scoresde,
	title        = {Score-Based Generative Modeling through Stochastic Differential Equations},
	author       = {Yang Song and Jascha Sohl-Dickstein and Diederik P Kingma and Abhishek Kumar and Stefano Ermon and Ben Poole},
	year         = 2021,
	booktitle    = {International Conference on Learning Representations}
}

@inproceedings{de2024diffusion,
	title        = {Diffusion Model Guided Sampling with Pixel-Wise Aleatoric Uncertainty Estimation},
	author       = {De Vita, Michele and Belagiannis, Vasileios},
	year         = 2025,
	month        = {February},
	booktitle    = {Proceedings of the Winter Conference on Applications of Computer Vision (WACV)},
	pages        = {3844--3854}
}

@article{xing2026divergence,
	title        = {Divergence is Uncertainty: A Closed-Form Posterior Covariance for Flow Matching},
	author       = {Xing, Jiarui and Wang, Song and Wang, Jian},
	year         = 2026,
	journal      = {arXiv preprint arXiv:2605.00941}
}

@article{albergo2025stochastic,
	title        = {Stochastic interpolants: A unifying framework for flows and diffusions},
	author       = {Albergo, Michael and Boffi, Nicholas M and Vanden-Eijnden, Eric},
	year         = 2025,
	journal      = {Journal of Machine Learning Research},
	volume       = 26,
	number       = 209,
	pages        = {1--80}
}

@inproceedings{black2024pi0,
	title        = {{$\pi_0$: A Vision-Language-Action Flow Model for General Robot Control}},
	author       = {Kevin Black AND Noah Brown AND Danny Driess AND Adnan Esmail AND Michael Robert Equi AND Chelsea Finn AND Niccolo Fusai AND Lachy Groom AND Karol Hausman AND Brian Ichter AND Szymon Jakubczak AND Tim Jones AND Liyiming Ke AND Sergey Levine AND Adrian Li-Bell AND Mohith Mothukuri AND Suraj Nair AND Karl Pertsch AND Lucy Xiaoyang Shi AND Laura Smith AND James Tanner AND Quan Vuong AND Anna Walling AND Haohuan Wang AND Ury Zhilinsky},
	year         = 2025,
	month        = {June},
	booktitle    = {Proceedings of Robotics: Science and Systems},
	address      = {LosAngeles, CA, USA},
	doi          = {10.15607/RSS.2025.XXI.010}
}

@article{bjorck2025gr00t,
	title        = {{GR00T N1} An open foundation model for generalist humanoid robots},
	author       = {Bjorck, Johan and Casta{\~n}eda, Fernando and Cherniadev, Nikita and Da, Xingye and Ding, Runyu and Fan, Linxi and Fang, Yu and Fox, Dieter and Hu, Fengyuan and Huang, Spencer and others},
	year         = 2025,
	journal      = {arXiv preprint arXiv:2503.14734}
}

@article{kendall2017uncertainties,
	title        = {What uncertainties do we need in bayesian deep learning for computer vision?},
	author       = {Kendall, Alex and Gal, Yarin},
	year         = 2017,
	journal      = {Advances in neural information processing systems},
	volume       = 30
}

@inproceedings{gal2016dropout,
	title        = {Dropout as a bayesian approximation: Representing model uncertainty in deep learning},
	author       = {Gal, Yarin and Ghahramani, Zoubin},
	year         = 2016,
	booktitle    = {international conference on machine learning},
	pages        = {1050--1059},
	organization = {PMLR}
}

@inproceedings{blundell2015weight,
	title        = {Weight uncertainty in neural network},
	author       = {Blundell, Charles and Cornebise, Julien and Kavukcuoglu, Koray and Wierstra, Daan},
	year         = 2015,
	booktitle    = {International conference on machine learning},
	pages        = {1613--1622},
	organization = {PMLR}
}

@inproceedings{pearce2020uncertainty,
	title        = {Uncertainty in neural networks: Approximately bayesian ensembling},
	author       = {Pearce, Tim and Leibfried, Felix and Brintrup, Alexandra},
	year         = 2020,
	booktitle    = {International conference on artificial intelligence and statistics},
	pages        = {234--244},
	organization = {PMLR}
}

@article{maddox2019simple,
	title        = {A simple baseline for bayesian uncertainty in deep learning},
	author       = {Maddox, Wesley J and Izmailov, Pavel and Garipov, Timur and Vetrov, Dmitry P and Wilson, Andrew Gordon},
	year         = 2019,
	journal      = {Advances in neural information processing systems},
	volume       = 32
}

@inproceedings{berry2024shedding,
	title        = {Shedding light on large generative networks: Estimating epistemic uncertainty in diffusion models},
	author       = {Berry, Lucas and Brando, Axel and Meger, David},
	year         = 2024,
	booktitle    = {The 40th Conference on Uncertainty in Artificial Intelligence}
}

@inproceedings{krueger2017hypernetworks,
	title        = {Bayesian Hypernetworks},
	author       = {David Krueger and Chin-Wei Huang and Riashat Islam and Ryan Turner and Alexandre Lacoste and Aaron Courville},
	year         = 2017,
	booktitle    = {Second Workshop on Bayesian Deep Learning at NeurIPS}
}

@article{chan2024estimating,
	title        = {Estimating epistemic and aleatoric uncertainty with a single model},
	author       = {Chan, Matthew A and Molina, Maria J and Metzler, Christopher A},
	year         = 2024,
	journal      = {Advances in neural information processing systems},
	volume       = 37,
	pages        = {109845--109870}
}

@inproceedings{jazbec2025generative,
	title        = {Generative Uncertainty in Diffusion Models},
	author       = {Metod Jazbec and Eliot Wong-Toi and Guoxuan Xia and Dan Zhang and Eric Nalisnick and Stephan Mandt},
	year         = 2025,
	booktitle    = {The 41st Conference on Uncertainty in Artificial Intelligence}
}

@inproceedings{gupta2026quantifying,
	title        = {Quantifying Epistemic Uncertainty in Diffusion Models},
	author       = {Aditi Gupta and Raphael A Meyer and Yotam Yaniv and Elynn Chen and N. Benjamin Erichson},
	year         = 2026,
	booktitle    = {The 29th International Conference on Artificial Intelligence and Statistics}
}

@article{de2026flowing,
	title        = {Flowing with Confidence},
	author       = {de Kruiff, Friso and Coscia, Dario and Welling, Max and Bekkers, Erik},
	year         = 2026,
	journal      = {arXiv preprint arXiv:2605.18472}
}

@inproceedings{mandlekar2021robomimic,
	title        = {What Matters in Learning from Offline Human Demonstrations for Robot Manipulation},
	author       = {Mandlekar, Ajay and Xu, Danfei and Wong, Josiah and Nasiriany, Soroush and Wang, Chen and Kulkarni, Rohun and Fei-Fei, Li and Savarese, Silvio and Zhu, Yuke and Mart\'in-Mart\'in, Roberto},
	year         = 2022,
	month        = {08--11 Nov},
	booktitle    = {Proceedings of the 5th Conference on Robot Learning},
	publisher    = {PMLR},
	series       = {Proceedings of Machine Learning Research},
	volume       = 164,
	pages        = {1678--1690}
}

@inproceedings{liu2023libero,
	title        = {{LIBERO}: Benchmarking Knowledge Transfer for Lifelong Robot Learning},
	author       = {Bo Liu and Yifeng Zhu and Chongkai Gao and Yihao Feng and qiang liu and Yuke Zhu and Peter Stone},
	year         = 2023,
	booktitle    = {Thirty-seventh Conference on Neural Information Processing Systems Datasets and Benchmarks Track}
}

@article{benton2024error,
	title        = {Error Bounds for Flow Matching Methods},
	author       = {Joe Benton and George Deligiannidis and Arnaud Doucet},
	year         = 2024,
	journal      = {Transactions on Machine Learning Research},
	issn         = {2835-8856}
}

@inproceedings{chen2023probability,
	title        = {The probability flow {ODE} is provably fast},
	author       = {Sitan Chen and Sinho Chewi and Holden Lee and Yuanzhi Li and Jianfeng Lu and Adil Salim},
	year         = 2023,
	booktitle    = {Thirty-seventh Conference on Neural Information Processing Systems}
}

@book{sarkka2013bayesian,
	title        = {Bayesian Filtering and Smoothing},
	author       = {S{\"a}rkk{\"a}, Simo},
	year         = 2013,
	publisher    = {Cambridge University Press},
	address      = {USA},
	isbn         = 1107619289
}

@book{hairer1993solving,
	title        = {Solving Ordinary Differential Equations {I} Nonstiff problems},
	author       = {Hairer, E. and N{\o}rsett, S.P. and Wanner, G.},
	year         = 2000,
	publisher    = {Springer},
	address      = {Berlin},
	edition      = {Second}
}

@article{shukor2025smolvla,
	title        = {{SmolVLA}: A Vision-Language-Action Model for Affordable and Efficient Robotics},
	author       = {Shukor, Mustafa and Aubakirova, Dana and Capuano, Francesco and Kooijmans, Pepijn and Palma, Steven and Zouitine, Adil and Aractingi, Michel and Pascal, Caroline and Russi, Martino and Marafioti, Andres and others},
	year         = 2025,
	journal      = {arXiv preprint arXiv:2506.01844}
}

@article{romer2025fiper,
	title        = {Failure Prediction at Runtime for Generative Robot Policies},
	author       = {Ralf R{\"o}mer and Adrian Kobras and Luca Worbis and Angela P. Schoellig},
	year         = 2025,
	journal      = {Advances in Neural Information Processing Systems (NeurIPS)}
}

@inproceedings{romer2026save,
	title        = {Uncertainty Quantification for Flow-Based Vision-Language-Action Models},
	author       = {Ralf R{\"o}mer and Maximilian Seeliger and Saida Liu and Ben Sturgis and Marco Bagatella and Daniel Marta and Andreas Krause and Angela P. Schoellig},
	year         = 2026,
	booktitle    = {RSS 2026 Workshop on Diffusion for Robot Learning}
}

@inproceedings{xu2025faildetect,
	title        = {{Can We Detect Failures Without Failure Data? Uncertainty-Aware Runtime Failure Detection for Imitation Learning Policies}},
	author       = {Chen Xu AND Tony Khuong Nguyen AND Emma Dixon AND Christopher Rodriguez AND Patrick Miller AND Robert Lee AND Paarth Shah AND Rares Andrei Ambrus AND Haruki Nishimura AND Masha Itkina},
	year         = 2025,
	month        = {June},
	booktitle    = {Proceedings of Robotics: Science and Systems},
	address      = {LosAngeles, CA, USA},
	doi          = {10.15607/RSS.2025.XXI.073}
}

@article{gu2025safe,
	title        = {Safe: Multitask failure detection for vision-language-action models},
	author       = {Gu, Qiao and Ju, Yuanliang and Sun, Shengxiang and Gilitschenski, Igor and Nishimura, Haruki and Itkina, Masha and Shkurti, Florian},
	year         = 2026,
	journal      = {Advances in Neural Information Processing Systems},
	volume       = 38,
	pages        = {40041--40076}
}

@inproceedings{agia2024sentinel,
	title        = {Unpacking Failure Modes of Generative Policies: Runtime Monitoring of Consistency and Progress},
	author       = {Christopher Agia and Rohan Sinha and Jingyun Yang and Ziang Cao and Rika Antonova and Marco Pavone and Jeannette Bohg},
	year         = 2024,
	booktitle    = {8th Annual Conference on Robot Learning},
}

@article{scheuerer2015variogram,
	title        = {Variogram-Based Proper Scoring Rules for Probabilistic Forecasts of Multivariate Quantities},
	author       = {Michael Scheuerer and Thomas M. Hamill},
	year         = 2015,
	journal      = {Monthly Weather Review},
	publisher    = {American Meteorological Society},
	volume       = 143,
	number       = 4,
	pages        = {1321--1334},
	doi          = {10.1175/MWR-D-14-00269.1}
}

@inproceedings{kristiadi2021lula,
	title        = {Learnable uncertainty under laplace approximations},
	author       = {Kristiadi, Agustinus and Hein, Matthias and Hennig, Philipp},
	year         = 2021,
	booktitle    = {Uncertainty in Artificial Intelligence},
	pages        = {344--353},
	organization = {PMLR}
}

@article{vanamersfoort2021feature,
	title        = {On Feature Collapse and Deep Kernel Learning for Single Forward Pass Uncertainty},
	author       = {van Amersfoort, Joost and Smith, Lewis and Jesson, Andrew and Key, Oscar and Gal, Yarin},
	year         = 2021,
	journal      = {arXiv preprint arXiv:2102.11409}
}

@inproceedings{mukhoti2023ddu,
	title        = {Deep Deterministic Uncertainty: A New Simple Baseline},
	author       = {Mukhoti, Jishnu and Kirsch, Andreas and van Amersfoort, Joost and Torr, Philip H.S. and Gal, Yarin},
	year         = 2023,
	month        = {June},
	booktitle    = {Proceedings of the IEEE/CVF Conference on Computer Vision and Pattern Recognition (CVPR)},
	pages        = {24384--24394}
}

@inproceedings{deng2022ella,
	title        = {Accelerated Linearized Laplace Approximation for Bayesian Deep Learning},
	author       = {Zhijie Deng and Feng Zhou and Jun Zhu},
	year         = 2022,
	booktitle    = {Advances in Neural Information Processing Systems},
	editor       = {Alice H. Oh and Alekh Agarwal and Danielle Belgrave and Kyunghyun Cho}
}

@inproceedings{calvo2026richer,
	title        = {Richer Bayesian Last Layers with Subsampled {NTK} Features},
	author       = {Sergio Calvo-Ordo{\~n}ez and Jonathan Plenk and Richard Bergna and Alvaro Cartea and Yarin Gal and Jos{\'e} Miguel Hern{\'a}ndez-Lobato and Kamil Ciosek},
	year         = 2026,
	booktitle    = {Forty-third International Conference on Machine Learning},
}

@article{jimenez2025psc,
	title        = {Probabilistic skip connections for deterministic uncertainty quantification in deep neural networks},
	author       = {Jimenez, Felix and Katzfuss, Matthias},
	year         = 2025,
	journal      = {arXiv preprint arXiv:2501.04816}
}

@inproceedings{he2016deep,
	title        = {Deep residual learning for image recognition},
	author       = {He, Kaiming and Zhang, Xiangyu and Ren, Shaoqing and Sun, Jian},
	year         = 2016,
	booktitle    = {Proceedings of the IEEE conference on computer vision and pattern recognition},
	pages        = {770--778}
}

@article{chi2023diffusion,
	title        = {Diffusion policy: Visuomotor policy learning via action diffusion},
	author       = {Chi, Cheng and Xu, Zhenjia and Feng, Siyuan and Cousineau, Eric and Du, Yilun and Burchfiel, Benjamin and Tedrake, Russ and Song, Shuran},
	year         = 2025,
	journal      = {The International Journal of Robotics Research},
	volume       = 44,
	number       = {10-11},
	pages        = {1684--1704}
}

@inproceedings{khazatsky2024droid,
	title        = {{DROID: A Large-Scale In-The-Wild Robot Manipulation Dataset}},
	author       = {Alexander Khazatsky AND Karl Pertsch AND Suraj Nair AND Ashwin Balakrishna AND Sudeep Dasari AND Siddharth Karamcheti AND Soroush Nasiriany AND Mohan Kumar Srirama AND Lawrence Yunliang Chen AND Kirsty Ellis AND Peter David Fagan AND Joey Hejna AND Masha Itkina AND Marion Lepert AND Yecheng Jason Ma AND Patrick Tree Miller AND Jimmy Wu AND Suneel Belkhale AND Shivin Dass AND Huy Ha AND Arhan Jain AND Abraham Lee AND Youngwoon Lee AND Marius Memmel AND Sungjae Park AND Ilija Radosavovic AND Kaiyuan Wang AND Albert Zhan AND Kevin Black AND Cheng Chi AND Kyle Beltran Hatch AND Shan Lin AND Jingpei Lu AND Jean Mercat AND Abdul Rehman AND Pannag R Sanketi AND Archit Sharma AND Cody Simpson AND Quan Vuong AND Homer Rich Walke AND Blake Wulfe AND Ted Xiao AND Jonathan Heewon Yang AND Arefeh Yavary AND Tony Z. Zhao AND Christopher Agia AND Rohan Baijal AND Mateo Guaman Castro AND Daphne Chen AND Qiuyu Chen AND Trinity Chung AND Jaimyn Drake AND Ethan Paul Foster AND Jensen Gao AND David Antonio Herrera AND Minho Heo AND Kyle Hsu AND Jiaheng Hu AND Donovon Jackson AND Charlotte Le AND Yunshuang Li AND Roy Lin AND Zehan Ma AND Abhiram Maddukuri AND Suvir Mirchandani AND Daniel Morton AND Tony Nguyen AND Abigail O'Neill AND Rosario Scalise AND Derick Seale AND Victor Son AND Stephen Tian AND Emi Tran AND Andrew E. Wang AND Yilin Wu AND Annie Xie AND Jingyun Yang AND Patrick Yin AND Yunchu Zhang AND Osbert Bastani AND Glen Berseth AND Jeannette Bohg AND Ken Goldberg AND Abhinav Gupta AND Abhishek Gupta AND Dinesh Jayaraman AND Joseph J Lim AND Jitendra Malik AND Roberto Martín-Martín AND Subramanian Ramamoorthy AND Dorsa Sadigh AND Shuran Song AND Jiajun Wu AND Michael C. Yip AND Yuke Zhu AND Thomas Kollar AND Sergey Levine AND Chelsea Finn},
	year         = 2024,
	month        = {July},
	booktitle    = {Proceedings of Robotics: Science and Systems},
	address      = {Delft, Netherlands},
	doi          = {10.15607/RSS.2024.XX.120}
}

@inproceedings{yang2026robolab,
	title        = {{RoboLab: A High-Fidelity Simulation Benchmark for Analysis of Task Generalist Policies}},
	author       = {Xuning Yang AND Rishit Dagli AND Alex Zook AND Hugo Hadfield AND Ankit Goyal AND Stan Birchfield AND Fabio Ramos AND Jonathan Tremblay},
	year         = 2026,
	month        = {July},
	booktitle    = {Proceedings of Robotics: Science and Systems},
	address      = {Sydney, Australia},
	doi          = {10.15607/RSS.2026.XXII.096}
}

@inproceedings{intelligence2025pi05,
	title        = {$\pi_{0.5}$: a Vision-Language-Action Model with Open-World Generalization},
	author       = {Black, Kevin and Brown, Noah and Darpinian, James and Dhabalia, Karan and Driess, Danny and Esmail, Adnan and Equi, Michael Robert and Finn, Chelsea and Fusai, Niccolo and Galliker, Manuel Y. and Ghosh, Dibya and Groom, Lachy and Hausman, Karol and ichter, brian and Jakubczak, Szymon and Jones, Tim and Ke, Liyiming and LeBlanc, Devin and Levine, Sergey and Li-Bell, Adrian and Mothukuri, Mohith and Nair, Suraj and Pertsch, Karl and Ren, Allen Z. and Shi, Lucy Xiaoyang and Smith, Laura and Springenberg, Jost Tobias and Stachowicz, Kyle and Tanner, James and Vuong, Quan and Walke, Homer and Walling, Anna and Wang, Haohuan and Yu, Lili and Zhilinsky, Ury},
	year         = 2025,
	month        = {27--30 Sep},
	booktitle    = {Proceedings of The 9th Conference on Robot Learning},
	publisher    = {PMLR},
	series       = {Proceedings of Machine Learning Research},
	volume       = 305,
	pages        = {17--40},
	editor       = {Lim, Joseph and Song, Shuran and Park, Hae-Won}
}

@article{beyer2024paligemma,
	title        = {{PaliGemma}: A versatile {3B} {VLM} for transfer},
	author       = {Beyer, Lucas and Steiner, Andreas and Pinto, Andr{\'e} Susano and Kolesnikov, Alexander and Wang, Xiao and Salz, Daniel and Neumann, Maxim and Alabdulmohsin, Ibrahim and Tschannen, Michael and Bugliarello, Emanuele and others},
	year         = 2024,
	journal      = {arXiv preprint arXiv:2407.07726}
}

@article{angelopoulos2023conformal,
	title        = {Conformal Prediction: A Gentle Introduction},
	author       = {Angelopoulos, Anastasios N. and Bates, Stephen},
	year         = 2023,
	journal      = {Found. Trends Mach. Learn.},
	publisher    = {Now Publishers Inc.},
	address      = {Hanover, MA, USA},
	volume       = 16,
	number       = 4,
	pages        = {494–591},
	doi          = {10.1561/2200000101},
	issn         = {1935-8237},
	issue_date   = {Mar 2023},
	numpages     = 114
}

@inproceedings{agia2025cupid,
	title        = {CUPID: Curating Data your Robot Loves with Influence Functions},
	author       = {Agia, Christopher and Sinha, Rohan and Yang, Jingyun and Antonova, Rika and Pavone, Marco and Nishimura, Haruki and Itkina, Masha and Bohg, Jeannette},
	year         = 2025,
	month        = {27--30 Sep},
	booktitle    = {Proceedings of The 9th Conference on Robot Learning},
	publisher    = {PMLR},
	series       = {Proceedings of Machine Learning Research},
	volume       = 305,
	pages        = {2907--2932},
	editor       = {Lim, Joseph and Song, Shuran and Park, Hae-Won}
}

@article{sirigiri2026diversity,
	title        = {Diversity You Can Actually Measure: A Fast, Model-Free Diversity Metric for Robotics Datasets},
	author       = {Sirigiri, Sreevardhan and de Lara, Nathan Samuel and Agia, Christopher and Shkurti, Florian and Ramos, Fabio},
	year         = 2026,
	journal      = {arXiv preprint arXiv:2603.11634}
}

@article{han2026flow,
	title        = {Flow Matching with Uncertainty Quantification and Guidance},
	author       = {Han, Juyeop and Beyer, Lukas Lao and Karaman, Sertac},
	year         = 2026,
	journal      = {arXiv preprint arXiv:2602.10326}
}

@article{berry2025emoe,
	title        = {EMoE: Training-Free Expert Disagreement for Uncertainty-Aware Text-to-Image Diffusion},
	author       = {Berry, Lucas and Brando, Axel and Chang, Wei-Di and Higuera, Juan Camilo Gamboa and Meger, David},
	year         = 2025,
	journal      = {arXiv preprint arXiv:2505.13273}
}

@inproceedings{bagatella2025active,
	title        = {Active Fine-Tuning of Multi-Task Policies},
	author       = {Marco Bagatella and Jonas H{\"u}botter and Georg Martius and Andreas Krause},
	year         = 2025,
	booktitle    = {Forty-second International Conference on Machine Learning}
}

@inproceedings{dass2026datamil,
	title        = {Data{MIL}: Selecting Data for Robot Imitation Learning with Datamodels},
	author       = {Shivin Dass and Alaa Khaddaj and Logan Engstrom and Aleksander Madry and Andrew Ilyas and Roberto Mart{\'\i}n-Mart{\'\i}n},
	year         = 2026,
	booktitle    = {The Fourteenth International Conference on Learning Representations}
}

@article{lakshminarayanan2017deep,
	title        = {Simple and scalable predictive uncertainty estimation using deep ensembles},
	author       = {Lakshminarayanan, Balaji and Pritzel, Alexander and Blundell, Charles},
	year         = 2017,
	journal      = {Advances in neural information processing systems},
	volume       = 30
}

@inproceedings{chen2023sampling,
	title        = {Sampling is as easy as learning the score: theory for diffusion models with minimal data assumptions},
	author       = {Sitan Chen and Sinho Chewi and Jerry Li and Yuanzhi Li and Adil Salim and Anru Zhang},
	year         = 2023,
	booktitle    = {The Eleventh International Conference on Learning Representations}
}

\newpage

\appendix 
\section{Appendix}
\addtocontents{toc}{\protect\setcounter{tocdepth}{2}}  
\appendixcontents

\subsection{Variance collapse}
\label{app:decay}

This appendix proves Theorem~\ref{thm:decay}. Proposition~\ref{prop:identity} shows that one denoising step of the recursion obeys the identity~\eqref{eq:identity}, Lemma~\ref{lem:transport} bounds the transport term in that identity, and the two combine into a covariance recursion that unrolls to the stated $O(1/T)$ upper bound.

\begin{proposition}[Trace identity for one denoising step]
\label{prop:identity}
Under the joint law of \S\ref{sec:setup}, the recursion~\eqref{eq:var_recursion} satisfies the identity~\eqref{eq:identity} at every step $i$ (re-stated here for convenience).
\end{proposition}
\begin{equation*}
\begin{aligned}
\Tr P_{i+1}
&= \Tr\Cov_x\!\left[x_i + \Delta t\, u_{\thetaMAP}(x_i, t_i)\right]
+ \Delta t^{2}\, \bar\sigma_i^{2} \\
&= \Tr\Cov_x\!\left[h_i(x_i)\right] + \Delta t^{2}\, \bar\sigma_i^{2}.
\end{aligned}
\tag{\ref{eq:identity}}
\end{equation*}

\begin{proof}
Taking the trace of~\eqref{eq:var_recursion} gives
\begin{equation}
\label{eq:summed}
\Tr P_{i+1} = \Tr P_i + 2\,\Delta t\, \Tr\Cov(x_i, u_i) + \Delta t^{2}\, \Tr\Cov(u_i),
\end{equation}
and both velocity terms are handled by conditioning on the state. The subscript $x$ denotes moments over $x_i$, omitting the step index for brevity.

\emph{The cross term.} Conditioning on the state, the law of total covariance splits it in two,
\begin{equation}
\label{eq:cross_step}
\Cov(x_i, u_i)
= \underbrace{\E_x\!\left[\Cov\!\left(x_i, u_i \mid x_i\right)\right]}_{=\;0}
+ \Cov_x\!\left(x_i,\, \E\!\left[u_i \mid x_i\right]\right)
= \Cov_x\!\left(x_i,\, u_{\thetaMAP}(x_i, t_i)\right),
\end{equation}
the first term vanishes because $x_i$ is deterministic given itself and the second collapses onto the MAP velocity under the joint law of \S\ref{sec:setup}, whose diagonal is exactly what the Monte Carlo estimator there targets.

\emph{The velocity variance.} Conditioning on the state again, the same law splits it in two as well,\footnote{BayesDiff uses conditional predictive variance~\citep[after Eq.~(8)]{kou2024bayesdiff}. We retain both terms of~\eqref{eq:var_step}; dropping the second, nonnegative term leaves our upper bound unchanged.}
\begin{equation}
\label{eq:var_step}
\Tr\Cov(u_i)
= \underbrace{\Tr\E_x\!\left[\Cov\!\left(u_i \mid x_i\right)\right]}_{=\;\bar\sigma_i^{2}}
+ \Tr\Cov_x\!\left(\E\!\left[u_i \mid x_i\right]\right)
= \bar\sigma_i^{2} + \Tr\Cov_x\!\left[u_{\thetaMAP}(x_i, t_i)\right],
\end{equation}
the first term averages the predictive covariance of~\eqref{eq:predictive} over the belief and the second measures the spread of the MAP velocity as $x_i$ varies under the current belief.

\emph{Recombining.} Substituting~\eqref{eq:cross_step} and~\eqref{eq:var_step} into~\eqref{eq:summed} leaves $\Delta t^{2}\bar\sigma_i^{2}$ alongside three terms involving the current state belief and MAP velocity, and bilinearity of the covariance collects those three into the Euler map $h_i(x_i) = x_i + \Delta t\, u_{\thetaMAP}(x_i, t_i)$ of \S\ref{sec:comp},
\begin{equation*}
\Tr P_i
+ 2\,\Delta t\, \Tr\Cov_x\!\left(x_i,\, u_{\thetaMAP}(x_i, t_i)\right)
+ \Delta t^{2}\, \Tr\Cov_x\!\left[u_{\thetaMAP}(x_i, t_i)\right]
= \Tr\Cov_x\!\left[h_i(x_i)\right],
\end{equation*}
which is the transport term of~\eqref{eq:identity}.
\end{proof}

\begin{lemma}[Transport barely grows the spread]
\label{lem:transport}
Under (A1) the transport term of~\eqref{eq:identity} obeys $\Tr\Cov_x[h_i(x_i)] \le (1 + \Delta t\, L)^{2}\, \Tr P_i$.
\end{lemma}

\begin{proof}
For any states $x_i,y_i \in \R^D$, the triangle inequality and (A1) give
\begin{equation*}
\left\|h_i(x_i) - h_i(y_i)\right\|
\le \left\|x_i - y_i\right\| + \Delta t\, \left\|u_{\thetaMAP}(x_i, t_i) - u_{\thetaMAP}(y_i, t_i)\right\|
\le \left(1 + \Delta t\, L\right) \left\|x_i - y_i\right\| .
\end{equation*}
Writing $m_i = \E_x[x_i]$, both $\E_x[h_i(x_i)]$ and $h_i(m_i)$ are deterministic quantities. Measuring squared distance from $h_i(m_i)$ instead of the transformed mean adds a nonnegative term:
\begin{equation*}
\E_x\!\left\|h_i(x_i) - h_i(m_i)\right\|^{2}
= \Tr\Cov_x[h_i(x_i)]
+ \left\|\E_x[h_i(x_i)] - h_i(m_i)\right\|^{2}.
\end{equation*}
Setting $y_i=m_i$ in the Lipschitz bound, squaring and taking expectations gives
\begin{equation*}
\begin{aligned}
\Tr\Cov_x[h_i(x_i)]
&\le \E_x\!\left\|h_i(x_i) - h_i(m_i)\right\|^{2} \\
&\le \left(1 + \Delta t\, L\right)^{2} \E_x\!\left\|x_i-m_i\right\|^{2}
= \left(1 + \Delta t\, L\right)^{2} \Tr P_i. \qedhere
\end{aligned}
\end{equation*}
\end{proof}

\begin{proof}[Proof of Theorem~\ref{thm:decay}]
Lemma~\ref{lem:transport} bounds the transport term of Proposition~\ref{prop:identity}, while (A2) bounds the injection by $\Delta t^{2}\bar\sigma_{\max}^{2}$, so~\eqref{eq:identity} is dominated by the scalar recursion
\begin{equation}
\label{eq:scalar_recursion}
\Tr P_{i+1} \le \alpha\, \Tr P_i + \beta,
\qquad
\alpha \triangleq (1 + \Delta t\, L)^{2} \ge 1,
\qquad
\beta \triangleq \Delta t^{2} \bar\sigma_{\max}^{2} .
\end{equation}
Unrolling~\eqref{eq:scalar_recursion} from $\Tr P_0 = 0$ leaves a geometric sum whose terms are each at most $\beta \alpha^{T}$, and substituting $\Delta t = 1/T$ gives:
\begin{equation*}
\Tr P_T
\;\le\; \beta \sum_{i=0}^{T-1} \alpha^{i}
\;\le\; T\, \beta\, \alpha^{T}
\;=\; \frac{\bar\sigma_{\max}^{2}}{T} \left(1 + \frac{L}{T}\right)^{\!2T}
\;\le\; e^{2L}\, \frac{\bar\sigma_{\max}^{2}}{T},
\end{equation*}
the last step because $1 + z \le e^{z}$ for every real $z$, which at $z = L/T$ raised to the $T$-th power gives $(1 + L/T)^{T} \le e^{L}$, and which is~\eqref{eq:decay}.
\end{proof}

\begin{corollary}[Full covariance]
\label{cor:fullcov}
Carrying the full covariance $P_{i+1} = \Cov(x_i + \Delta t\, u_i)$ in place of the diagonal of~\eqref{eq:var_recursion} leaves Theorem~\ref{thm:decay} unchanged.
\end{corollary}

\begin{proof}
Expanding the full covariance of the Euler update~\eqref{eq:euler} gives,
\begin{equation}
\label{eq:fullcov}
P_{i+1} = P_i + \Delta t\, \big[\Cov(x_i, u_i) + \Cov(u_i, x_i)\big] + \Delta t^{2}\, \Cov(u_i),
\end{equation}
and taking the trace, with $\Tr\Cov(u_i, x_i) = \Tr\Cov(x_i, u_i)^{\transpose} = \Tr\Cov(x_i, u_i)$,
\begin{equation*}
\Tr P_{i+1} = \Tr P_i + 2\,\Delta t\, \Tr\Cov(x_i, u_i) + \Delta t^{2}\, \Tr\Cov(u_i),
\end{equation*}
which is~\eqref{eq:summed}. Lemma~\ref{lem:transport} holds and~\eqref{eq:scalar_recursion} unrolls as before. 
\end{proof}

\begin{corollary}[Diffusion samplers]
\label{cor:sde}
Let time run from noise at $t = 0$ to data at $t = 1$ as in \S\ref{sec:setup}. The reverse-time SDE through which BayesDiff propagates uncertainty~\citep{song2021scoresde, kou2024bayesdiff} is
\begin{equation}
\label{eq:reverse_sde}
\dd x_t = b_\theta(x_t, t)\, \dd t + g(t)\, \dd\omega_t,
\qquad
b_\theta(x, t) \triangleq -\Big[f(t)\, x + \frac{g(t)^{2}}{\sigma_t}\, \epsilon_\theta(x, t)\Big],
\end{equation}%

where $f$ and $g$ are the drift and diffusion coefficients respectively of the forward noising process, $\sigma_t$ is the standard deviation of the noise at time $t$, $\omega_t$ is a Wiener process, and $\epsilon_\theta$ is the noise prediction network, over whose last layer the Laplace posterior~\eqref{eq:llla} is placed. The drift $b_\theta$ plays the role of the velocity field in~\eqref{eq:euler} and is affine in $\epsilon_\theta$, while the diffusion term $g(t)\, \dd\omega_t$ does not depend on $\theta$. We take $f$, $g$ and $g^{2} / \sigma_t$ bounded on $[0, 1]$, which holds when the sampled interval ends at a fixed positive distance from the point $\sigma_t = 0$, as done in continuous-time sampling~\citep{song2021scoresde}. Using the same integration steps as~\eqref{eq:euler}, the Euler-Maruyama discretization of~\eqref{eq:reverse_sde} gives
\begin{equation}
\label{eq:em_step}
x_{i+1} = x_i + \Delta t\, b_\theta(x_i, t_i) + \sqrt{\Delta t}\, g_i\, \eta_i,
\qquad
g_i \triangleq g(t_i),
\qquad
\eta_i \sim \Normal(0, I),
\end{equation}
with $\eta_i$ independent of $x_i$ and $\theta$, and the iteration rule of BayesDiff is the moment recursion~\eqref{eq:mean_recursion}--\eqref{eq:var_recursion} of this step, written with the step size absorbed into the coefficients~\citep[Eq.~7]{kou2024bayesdiff}. Under (A1) for $\epsilon_{\thetaMAP}$, analogous to the Lipschitz assumption on the true score in~\citep[Assumption~1]{chen2023sampling}, the MAP drift $b_{\thetaMAP}$ is Lipschitz. Under (A2) for the noise predictor variance and the bounded schedule coefficients, the summed predictive variance of $b_\theta$ is also bounded. We again denote the resulting constants by $L$ and $\bar\sigma_{\max}^{2}$, and the recursion reports
\begin{equation}
\label{eq:sde_bound}
\Tr P_T \;\le\; e^{2L} \Bigg( \underbrace{\frac{\bar\sigma_{\max}^{2}}{T}}_{\text{epistemic}} + \underbrace{D\, \Delta t \sum_{i=0}^{T-1} g_i^{2}}_{\text{aleatoric}} \Bigg),
\end{equation}
so the epistemic term in the upper bound decays as in Theorem~\ref{thm:decay}, while the aleatoric term is $O(1)$.
\end{corollary}

\begin{proof}
Conditioned on $x_i$, the noise prediction has the Gaussian form of~\eqref{eq:predictive}. In the drift~\eqref{eq:reverse_sde}, the term $-f(t_i)x_i$ is deterministic, so with $\sigma_i \triangleq \sigma_{t_i}$,
\begin{equation*}
\Cov\!\left[b_\theta(x_i,t_i)\mid x_i\right]
= \left(\frac{g_i^2}{\sigma_i}\right)^2
  \Cov\!\left[\epsilon_\theta(x_i,t_i)\mid x_i\right]
= \frac{g_i^4}{\sigma_i^2}\,\diag\gamma^2(x_i,t_i).
\end{equation*}
The drift is affine in the noise prediction, so its predictive distribution is Gaussian as in~\eqref{eq:predictive}, with mean $b_{\thetaMAP}(x_i,t_i)$:
\begin{equation}
\label{eq:drift_predictive}
p\!\left(b \mid x_i, t_i, \D\right) \approx \Normal\!\Big(b;\, b_{\thetaMAP}(x_i, t_i),\, \frac{g_i^{4}}{\sigma_i^{2}}\, \diag \gamma^{2}(x_i, t_i)\Big).
\end{equation}
Without the Brownian increment, the step~\eqref{eq:em_step} is the Euler step~\eqref{eq:euler} with $b_\theta$ in place of $u_\theta$. Proposition~\ref{prop:identity} therefore applies with~\eqref{eq:drift_predictive} in place of~\eqref{eq:predictive}, and Lemma~\ref{lem:transport} bounds the transport term,
\begin{equation*}
\begin{aligned}
\Tr\Cov\!\left[x_i + \Delta t\, b_\theta(x_i, t_i)\right]
&= \Tr\Cov_x\!\left[x_i + \Delta t\, b_{\thetaMAP}(x_i, t_i)\right] + \Delta t^{2}\, \bar\sigma_i^{2} \\
&\le (1 + \Delta t\, L)^{2}\, \Tr P_i + \Delta t^{2}\, \bar\sigma_{\max}^{2},
\end{aligned}
\end{equation*}
where $\bar\sigma_i^{2} \triangleq \E_x\big[\tfrac{g_i^{4}}{\sigma_i^{2}} \sum_{d=1}^{D} \gamma^{2}_d(x_i, t_i)\big]$ is the variance of~\eqref{eq:drift_predictive} summed over coordinates and averaged over the belief, the counterpart of the $\bar\sigma_i^{2}$ of~\eqref{eq:identity}, which is bounded by $\bar\sigma_{\max}^{2}$. Since the aleatoric increment $\sqrt{\Delta t}\, g_i \eta_i$ is zero-mean and independent of $(x_i, \theta)$, it adds its covariance without a cross term, $P_{i+1} = \Cov\!\left[x_i + \Delta t\, b_\theta(x_i, t_i)\right] + g_i^{2} \Delta t\, I$, and~\eqref{eq:identity} becomes
\begin{equation*}
\Tr P_{i+1}
= \Tr\Cov_x\!\left[x_i + \Delta t\, b_{\thetaMAP}(x_i, t_i)\right]
+ \Delta t^{2}\, \bar\sigma_i^{2}
+ D\, g_i^{2}\, \Delta t
\;\le\; \alpha\, \Tr P_i + \beta + D\, g_i^{2}\, \Delta t,
\end{equation*}

with $\alpha$ and $\beta$ as in~\eqref{eq:scalar_recursion}. Unrolling from $\Tr P_0 = 0$ as in the proof of Theorem~\ref{thm:decay}, with every power of $\alpha$ bounded by $\alpha^{T} \le e^{2L}$, gives
\begin{equation*}
\Tr P_T
\;\le\; e^{2L} \sum_{i=0}^{T-1} \left(\beta + D\, g_i^{2}\, \Delta t\right)
\;=\; e^{2L} \Bigg(\frac{\bar\sigma_{\max}^{2}}{T} + D\, \Delta t \sum_{i=0}^{T-1} g_i^{2}\Bigg),
\end{equation*}
which is~\eqref{eq:sde_bound}. The two sources of randomness enter~\eqref{eq:em_step} at different orders in the step size: the weight deviation through $\Delta t\, b_\theta$, with a variance of order $\Delta t^{2}$ per step and an $O(1/T)$ term in the bound, and the Brownian increment through $\sqrt{\Delta t}\, g_i \eta_i$, with a variance of order $\Delta t$ per step and an $O(1)$ term in the bound. The ancestral sampler of DDPM~\citep{ho2020ddpm} is a discretization of this form for the variance-preserving schedule~\citep{song2021scoresde}, and deterministic samplers of the probability-flow ODE drop the increment, halve the score term, and inherit the bound of Theorem~\ref{thm:decay} for their drift.
\end{proof}

\subsection{Upper bound independent of integration steps}
\label{app:coherence}

This appendix proves Theorem~\ref{thm:stable}. Proposition~\ref{prop:unroll} unrolls the recursion into a sum over the integration steps, and bounding that sum under (A1) and (A2) gives the two upper bounds. 

\begin{proposition}[The chain unrolls into one sum]
\label{prop:unroll}
Run from $(x_0, 0, 0)$, the recursion~\eqref{eq:comp_mean}--\eqref{eq:comp_state} reports
\begin{equation}
\label{eq:endpoint}
P_T = M_T\, \Sigma_\theta\, M_T^{\transpose},
\qquad
M_T \triangleq \Delta t \sum_{i=0}^{T-1} \Phi_i\, J^{\theta}_i,
\qquad
\Phi_i \triangleq A_{T-1} \cdots A_{i+1},
\end{equation}
with  $\Phi_{T-1} = I$, while the same recursion with $C_i$ held at zero reports
\begin{equation}
\label{eq:endpoint_zero}
P_T = \Delta t^{2} \sum_{i=0}^{T-1} \Phi_i\, J^{\theta}_i\, \Sigma_\theta\, J^{\theta\transpose}_i\, \Phi_i^{\transpose} .
\end{equation}
\end{proposition}

\begin{proof}
Recall that $z_i = (\delta x_i,\delta\theta)$ stacks the state deviation $\delta x_i=x_i-m_i$ and the fixed weight deviation. Its covariance and transition matrix are
\begin{equation*}
\Cov(z_i) =
\begin{pmatrix}
P_i & C_i \\
C_i^{\transpose} & \Sigma_\theta
\end{pmatrix},
\qquad
F_i =
\begin{pmatrix}
A_i & B_i \\
0 & I
\end{pmatrix}.
\end{equation*}
The upper-left block $P_i=\Cov(\delta x_i)$ is the state covariance, and the upper-right block $C_i=\Cov(\delta x_i,\delta\theta)$ captures its dependence on the weights. The joint update $\Cov(z_{i+1})=F_i\Cov(z_i)F_i^{\transpose}$ gives~\eqref{eq:comp_cross}--\eqref{eq:comp_state} for these two blocks. Starting from $P_0=C_0=0$, we unroll the recursion with and without the cross-covariance below.

\emph{With the cross-covariance.} Applying~\eqref{eq:comp_lin} from $\delta x_0 = 0$ injects $B_i \delta\theta$ at step $i$ and multiplies it by every $A$ that follows, the same $\delta\theta$ appearing in all $T$ of them, so the endpoint collects
\begin{equation*}
\delta x_T
= \sum_{i=0}^{T-1} A_{T-1} \cdots A_{i+1}\, B_i\, \delta\theta
= \Delta t \sum_{i=0}^{T-1} \Phi_i\, J^{\theta}_i\, \delta\theta
= M_T\, \delta\theta,
\end{equation*}
using $B_i = \Delta t\, J^{\theta}_i$, and since $\delta\theta \sim \Normal(0, \Sigma_\theta)$ its covariance is~\eqref{eq:endpoint}.

\emph{Without it.} Holding $C_i \equiv 0$ leaves $P_{i+1} = A_i P_i A_i^{\transpose} + B_i \Sigma_\theta B_i^{\transpose}$, the transported state covariance and a weight contribution injected afresh, with no term relating the two. The recursion~\eqref{eq:var_recursion} of \S\ref{sec:setup} carries no such term either, the weights having been marginalized out of the velocity field~\eqref{eq:predictive} before the step is taken, and we attribute the decay of Theorem~\ref{thm:decay} to that omission. Particularly, \citet{kou2024bayesdiff} assume $\Cov(x_s, \epsilon_t) = 0$ for $s \neq t$, where $\epsilon_t$ is their noise prediction (see their Appendix~A.2, after Eq.~32). Unrolling from $P_0 = 0$ transports the injection $B_i \Sigma_\theta B_i^{\transpose}$ through the same steps that follow,
\begin{equation*}
P_T
= \sum_{i=0}^{T-1} \Phi_i\, B_i \Sigma_\theta B_i^{\transpose}\, \Phi_i^{\transpose}
= \Delta t^{2} \sum_{i=0}^{T-1} \Phi_i\, J^{\theta}_i\, \Sigma_\theta\, J^{\theta\transpose}_i\, \Phi_i^{\transpose},
\end{equation*}
which is~\eqref{eq:endpoint_zero}, the identical $T$ contributions with their cross terms removed.
\end{proof}

\begin{proof}[Proof of Theorem~\ref{thm:stable}]
Both upper bounds follow from the same two bounds on a single summand. Under (A1) each factor of the transport obeys $\|A_i\| \le 1 + \Delta t\, L$, so the product of at most $T$ of them obeys $\|\Phi_i\| \le (1 + \Delta t\, L)^{T} \le e^{L}$. Under (A2), writing $\Sigma_\theta^{1/2}$ for the symmetric square root of $\Sigma_\theta$ gives
\begin{equation*}
\|J^{\theta}_i \Sigma_\theta^{1/2}\|_F^2
= \Tr[J^{\theta}_i \Sigma_\theta J^{\theta\transpose}_i]
= \sum_d \gamma_d^2(m_i,t_i)
\le \bar\sigma_{\max}^2.
\end{equation*}
Taking square roots yields $\|J^{\theta}_i \Sigma_\theta^{1/2}\|_F \le \bar\sigma_{\max}$. Here $\|\cdot\|$ denotes the spectral norm and $\|\cdot\|_F$ the Frobenius norm. Taking the trace of~\eqref{eq:endpoint} gives $\Tr P_T = \|M_T\Sigma_\theta^{1/2}\|_F^2$. Substituting $M_T = \Delta t\sum_i \Phi_i J_i^\theta$, the triangle inequality bounds the norm of the sum by the sum of the norms. We then apply $\|AB\|_F \le \|A\|\,\|B\|_F$ to each term, separating $\Phi_i$ from $J_i^\theta\Sigma_\theta^{1/2}$, and use the bounds established above:
\begin{align*}
\Tr P_T
&= \left\|M_T\, \Sigma_\theta^{1/2}\right\|_{F}^{2}
\;=\; \Delta t^{2}\left\|\sum_{i=0}^{T-1} \Phi_i\, J^{\theta}_i\, \Sigma_\theta^{1/2}\right\|_{F}^{2} \\[2pt]
&\le\; \left(\Delta t \sum_{i=0}^{T-1} \left\|\Phi_i\, J^{\theta}_i\, \Sigma_\theta^{1/2}\right\|_{F}\right)^{\!2}
\;\le\; \left(\Delta t \sum_{i=0}^{T-1}
\underbrace{\left\|\Phi_i\right\|}_{\le\; e^{L}}\;
\underbrace{\left\|J^{\theta}_i\, \Sigma_\theta^{1/2}\right\|_{F}}_{\le\;\bar\sigma_{\max}}\right)^{\!2} \\[2pt]
&\le\; \left(\Delta t\, T\, e^{L}\, \bar\sigma_{\max}\right)^{\!2}
\;=\; e^{2L}\, \bar\sigma_{\max}^{2},
\end{align*}
where the last equality uses $\Delta t\,T=1$, giving~\eqref{eq:stable}.

Taking the trace of~\eqref{eq:endpoint_zero} gives a sum of squared Frobenius norms. Applying the same bounds to each term,
\begin{align*}
\Tr P_T
&= \Delta t^{2} \sum_{i=0}^{T-1}
\|\Phi_i J_i^\theta \Sigma_\theta^{1/2}\|_F^{2} 
\le \Delta t^{2} \sum_{i=0}^{T-1}
\|\Phi_i\|^{2}\,
\|J_i^\theta \Sigma_\theta^{1/2}\|_F^{2} \\
&\le T\,\Delta t^{2}\,e^{2L}\bar\sigma_{\max}^{2}
= e^{2L}\frac{\bar\sigma_{\max}^{2}}{T},
\end{align*}
where $T\Delta t^{2}=1/T$, giving~\eqref{eq:decay}.

\end{proof}

\subsection{Supplementary Details: Augmented Head}
\label{app:cond_head}
For a linear last layer under the flow matching loss, the curvature in~\eqref{eq:llla} is the Gram matrix of the last-layer input over the fit set $\D_{\mathrm{fit}}$ plus $\lambda I$, so widening the input from $\phi$ to $\psi$ of~\eqref{eq:cond_head} changes the Laplace posterior while preserving the policy's action predictions. Here, $\zeta = c(o)$ contains the observation embeddings already computed by the policy. Writing $\Sigma_d$ for the block of $\Sigma_\theta$ that belongs to the $d$-th row of $\theta$, the variance~\eqref{eq:predictive} of the $d$-th velocity coordinate expands as
\begin{equation}
\label{eq:cond_variance}
\gamma^{2}_d(x, t) = \psi^{\transpose} \Sigma_d\, \psi
= \begin{pmatrix} \phi \\ \zeta \end{pmatrix}^{\!\transpose}
\begin{pmatrix} \Sigma_d^{\phi\phi} & \Sigma_d^{\phi\zeta} \\ \Sigma_d^{\zeta\phi} & \Sigma_d^{\zeta\zeta} \end{pmatrix}
\begin{pmatrix} \phi \\ \zeta \end{pmatrix}
= \phi^{\transpose} \Sigma_d^{\phi\phi}\, \phi + 2\, \phi^{\transpose} \Sigma_d^{\phi\zeta}\, \zeta + \zeta^{\transpose} \Sigma_d^{\zeta\zeta}\, \zeta,
\end{equation}%
\noindent
where the first term uses the original features under the augmented posterior; the remaining terms involve the appended observation embeddings. Write $\theta_d = (\theta_d^\phi,\theta_d^\zeta)$, separating the weights on the original features $\phi$ from those on the appended observation embeddings $\zeta$. To compare the two Laplace heads' predictive variances, we recover the plain head by fixing the appended weights $\theta_d^\zeta$ to zero. With the same fit data and prior, its weight covariance is the conditional covariance under the augmented posterior:
\begin{equation}
\label{eq:cond_precision}
\Cov(\theta_d^\phi \mid \theta_d^\zeta=0)
= (\Lambda_d^{\phi\phi})^{-1},
\qquad
\Lambda_d \triangleq \Sigma_d^{-1},
\end{equation}%
\noindent where $\Lambda_d^{\phi\phi}$ is the Gram matrix of $\phi$ plus $\lambda I$, which is the precision of the plain Laplace head, and the conditional covariance of a Gaussian does not depend on the value conditioned on. The law of total variance then gives
\begin{equation}
\label{eq:cond_bound}
\begin{aligned}
\psi^{\transpose} \Sigma_d\, \psi
&= \E\big[\operatorname{Var}(\psi^{\transpose}\theta_d \mid \theta_d^{\zeta})\big] + \operatorname{Var}\big(\E[\psi^{\transpose}\theta_d \mid \theta_d^{\zeta}]\big) \\
&= \phi^{\transpose} (\Lambda_d^{\phi\phi})^{-1} \phi + \operatorname{Var}\big(\E[\psi^{\transpose}\theta_d \mid \theta_d^{\zeta}]\big)
\;\ge\; \phi^{\transpose} (\Lambda_d^{\phi\phi})^{-1} \phi,
\end{aligned}
\end{equation}%
\noindent
where the right-hand side is the variance~\eqref{eq:predictive} of the plain head, so the augmented head's predictive velocity variance is no smaller than the plain head's at any input. Proposition~2 of \citet{kristiadi2021lula} states this for a diagonal Laplace, and the argument holds equally for the Kronecker-factored posterior we use, whose input factor is the Gram matrix of $\psi$. \rebuttal{Note that for a linear last layer, the Gram matrix depends only on the features and not on $\theta$, so the covariance obtained at $\thetaMAPbar$ equals the one at the mode of the augmented posterior, and only the mean differs.} Like LULA~\citep{kristiadi2021lula}, our augmented head aims to alleviate feature collapse by widening the last-layer features, here using existing observation embeddings. Related approaches enrich Bayesian last layers with projected neural tangent kernel features~\citep{calvo2026richer} or fit probabilistic predictors to intermediate representations~\citep{jimenez2025psc}.\looseness-1

\subsection{Supplementary Details: Pseudocode}
\label{app:algo_ekf}

Algorithms~\ref{alg:ekf} and~\ref{alg:ukf} list the CoMP-EKF and CoMP-UKF recursions of \S\ref{sec:comp} step by step; they differ in the propagated mean, transition $A_i$ and transported state covariance.

\paragraph{CoMP-UKF moment updates.}

The state transition $A_i$ in~\eqref{eq:comp_lin} and transported spread $A_i P_i A_i^{\transpose}$ in~\eqref{eq:comp_state} can also be estimated from deterministic points instead of derivatives. The unscented transform~\citep{sarkka2013bayesian} places $2D+1$ sigma points $\chi_j$ around $m_i$ along the columns of a square root of $P_i$, with mean weights $w_j^{(m)}$ and covariance weights $w_j^{(c)}$ chosen so that the weighted points carry the mean and the covariance of the belief. Pushing them through the map, $y_j = h_i(\chi_j)$ with $\bar y \triangleq \sum_j w_j^{(m)} y_j$, gives 
\begin{equation}
\label{eq:comp_ukf}
\Sigma^{yy}_i = \sum\nolimits_j w_j^{(c)}\, (y_j - \bar y)(y_j - \bar y)^{\transpose},
\qquad
\Sigma^{xy}_i = \sum\nolimits_j w_j^{(c)}\, (\chi_j - m_i)(y_j - \bar y)^{\transpose},
\end{equation}
\noindent
which estimate the covariance of the transported state and its covariance with $x_i$. Two substitutions close the recursion: the transported spread $A_i P_i A_i^{\transpose}$ in~\eqref{eq:comp_state} becomes $\Sigma^{yy}_i$, retaining curvature beyond the linear approximation, while elsewhere we use $A_i = (\Sigma^{xy}_i)^{\transpose} P_i^{-1}$, the least-squares affine surrogate of $h_i$ over the current belief~\citep{sarkka2013bayesian}. The mean advances as $m_{i+1} = \bar y$ and the weight terms are computed analytically using $J^{\theta}_i$ evaluated at $m_i$, so no sigma points are needed for $\theta$. Algorithm~\ref{alg:ukf} computes $A_i$ by solving the regularized system $(P_i + \varepsilon_i I) A_i^{\transpose} = \Sigma^{xy}_i$, where $\varepsilon_i > 0$ is a small regularization parameter that keeps the solve defined even at $P_0 = 0$.

\begin{algo}{Coherent Moment Propagation with the extended Kalman filter (CoMP-EKF)}
\label{alg:ekf}
\noindent
\textbf{Require:} sample $x_0 \sim \Normal(0, I)$, steps $T$ with $\Delta t = 1/T$, MAP velocity $u_{\thetaMAP}$ with features $\phi$, last-layer posterior $\Sigma_\theta$ \\
\textbf{Output:} sample $m_T$ and covariance $P_T$
\begin{algosteps}
  \item Initialize $m_0 \leftarrow x_0$, $P_0 \leftarrow \bm{0}$, $C_0 \leftarrow \bm{0}$
  \item \textbf{for} $i = 0 \to T-1$ \textbf{do}
  \item \quad $J^{x}_i \leftarrow \partial u_{\thetaMAP}(x_i, t_i) / \partial x_i$ at $x_i = m_i$
  \item \quad $A_i \leftarrow I + \Delta t\, J^{x}_i$ \quad and \quad $B_i \leftarrow \Delta t\, J^{\theta}_i$ \quad \textcolor{black!60}{(analytic $J^{\theta}_i$ from $\phi$ for linear last-layer)}
  \item \quad $m_{i+1} \leftarrow m_i + \Delta t\, u_{\thetaMAP}(m_i, t_i)$
  \item \quad $C_{i+1} \leftarrow A_i C_i + B_i \Sigma_\theta$
  \item \quad $P_{i+1} \leftarrow A_i P_i A_i^{\transpose} + A_i C_i B_i^{\transpose} + B_i C_i^{\transpose} A_i^{\transpose} + B_i \Sigma_\theta B_i^{\transpose}$
  \item \textbf{end for}
  \item \textbf{return} $\mathcal{N}(m_T, P_T)$
\end{algosteps}
\end{algo}

\begin{algo}{Coherent Moment Propagation with the unscented Kalman filter (CoMP-UKF)}
\label{alg:ukf}
\noindent
\textbf{Require:} sample $x_0 \sim \Normal(0, I)$, steps $T$ with $\Delta t = 1/T$, MAP velocity $u_{\thetaMAP}$ with features $\phi$, last-layer posterior $\Sigma_\theta$ \\
\textbf{Output:} sample $m_T$ and covariance $P_T$
\begin{algosteps}
  \item Initialize $m_0 \leftarrow x_0$, $P_0 \leftarrow \bm{0}$, $C_0 \leftarrow \bm{0}$
  \item \textbf{for} $i = 0 \to T-1$ \textbf{do}
  \item \quad Compute the $2D+1$ sigma points $\chi_j$ and weights $w_j^{(m)}, w_j^{(c)}$ carrying $(m_i, P_i)$ by the unscented transform
  \item \quad $y_j \leftarrow \chi_j + \Delta t\, u_{\thetaMAP}(\chi_j, t_i)$ \quad and \quad $\bar y \leftarrow \sum_j w_j^{(m)} y_j$
  \item \quad $\Sigma^{yy}_i$ and $\Sigma^{xy}_i$ from~\eqref{eq:comp_ukf}, then solve $(P_i + \varepsilon_i I) A_i^{\transpose} = \Sigma^{xy}_i$ for $A_i$ and $B_i \leftarrow \Delta t\, J^{\theta}_i$ \quad \textcolor{black!60}{(closed-form $B_i$)}
  \item \quad $m_{i+1} \leftarrow \bar y$
  \item \quad $C_{i+1} \leftarrow A_i C_i + B_i \Sigma_\theta$
  \item \quad $P_{i+1} \leftarrow \Sigma^{yy}_i + A_i C_i B_i^{\transpose} + B_i C_i^{\transpose} A_i^{\transpose} + B_i \Sigma_\theta B_i^{\transpose}$
  \item \textbf{end for}
  \item \textbf{return} $\mathcal{N}(m_T, P_T)$
\end{algosteps}
\noindent
\end{algo}

\FloatBarrier

\subsection{Supplementary Details: Uncertainty Scores}
\label{app:score}

Algorithms~\ref{alg:ekf} and~\ref{alg:ukf} return the full covariance $P_T$ of the predicted action chunk~\eqref{eq:chunk}. We summarize it with two scalar uncertainty scores:
\begin{equation}
\label{eq:score}
U_{\mathrm{chunk}} \triangleq \Tr P_T,
\qquad
U_{\mathrm{endpoint}} \triangleq \Tr\Cov\Bigg(\sum_{k=1}^{H} a_k\Bigg),
\end{equation}
\noindent
where $H$ is the number of actions in the predicted chunk. The chunk score gives the total variance, summing the variances of all action coordinates. In contrast, the endpoint score also includes contributions from the off-diagonal blocks of $P_T$ that couple commands at different steps.

The covariance of the accumulated commands can be written in terms of the blocks of $P_T$ as
\begin{equation*}
\Sigma_H \triangleq \Cov\Bigg(\sum_{k=1}^{H} a_k\Bigg)
= \sum_{k=1}^{H} \sum_{k'=1}^{H} [P_T]_{k k'},
\qquad
U_{\mathrm{endpoint}} = \Tr\Sigma_H,
\end{equation*}
\noindent
where $[P_T]_{k k'}$ is the covariance block coupling actions $k$ and $k'$. Summing the commands treats the chunk as one accumulated action, whose uncertainty depends on both the individual command variances and their covariance across steps. We use $U = U_{\mathrm{chunk}}$ for direct comparison with BayesDiff in \S\ref{sec:calibration} and $U = U_{\mathrm{endpoint}}$ in all manipulation applications (\S\ref{sec:conditioning}, \S\ref{sec:failure} and \S\ref{sec:finetuning}).

\newpage

\subsection{Experimental Settings}
\label{app:implementation}

Table~\ref{tab:hparams} summarizes the settings used for the baselines and our methods across experiments. Experiment-specific fitting, sampling and evaluation protocols are given in Appendices~\ref{app:faithfulness}--\ref{app:oneshot}.

\begin{table}[!htpb]
  \centering
  \caption{\textbf{Settings of the uncertainty estimators.} Baseline settings follow the released implementations, with adaptations to flow policies and experiment-specific sampling budgets. For each policy, we evaluate the post-hoc estimators with both the plain and augmented last-layer Laplace posteriors. $K$ counts candidate chunks from independent initial noises per observation, $M$ posterior weight draws (trained members for ensemble references), and $S$ state particles; sampling budgets are given with each experiment.}
  \label{tab:hparams}
  \small
  \setlength{\tabcolsep}{4pt}
  \renewcommand{\arraystretch}{1.12}
  \begin{tabular}{@{}>{\raggedright\arraybackslash}p{0.16\linewidth}>{\raggedright\arraybackslash}p{0.54\linewidth}>{\raggedright\arraybackslash}p{\dimexpr0.30\linewidth-4\tabcolsep\relax}@{}}
  \toprule
  \textbf{Estimator} & \textbf{Settings} & \textbf{Score $U$} \\
  \midrule
  \multicolumn{3}{@{}l}{\emph{Common setup for post-hoc estimators}} \\
  Posterior & Kronecker-factored Laplace of the last layer, prior precision $\lambda = 1$; plain head on $\phi$, augmented head on $\psi = (\phi, c(o))$; $\Sigma_\theta$ fixed during integration & \\
  Sampler & Euler integration with $K = 1$ except for VFD; step counts, ablations and VFD candidate budgets are specified per experiment & \\
  \midrule
  BayesDiff \citep{kou2024bayesdiff} & Diagonal moment recursion~\eqref{eq:mean_recursion}--\eqref{eq:var_recursion}; state-velocity cross-covariance estimated from $S$ state particles per step & $\Tr P_T$ \\
  DEUA \citep{huang2026deua} & MAP sample re-noised to $\tau = 0.98$, one moment step to $\tau = 0.99$ ($\Delta t = 0.01$); velocity variance over the $M$ draws at that state & $\Delta t^{2} \sum_d \operatorname{Var}_\theta[u_d]$ \\
  GU \citep{jazbec2025generative} & $M$ draws plus the MAP member integrated to $t = 1$; Gaussian entropy of the per-coordinate variance of the terminal state, floor $\sigma_0^{2} = 10^{-8}$ & $\begin{aligned}[t] &\textstyle \tfrac12 \sum_d \log(\mathrm{var}_d + \sigma_0^{2}) \\ &\textstyle \quad + \tfrac D2\bigl[1 + \log(2\pi)\bigr] \end{aligned}$ \\  DECU \citep{berry2024shedding} & Flow adaptation following SAVE~\citep{romer2026save}: shared MAP trunk to the branching time $t = 0.995$, then one velocity $u^{(i)}$ for each of the $M$ members; PaiDE mutual-information proxy with $D_{ij} = \|u^{(i)} - u^{(j)}\|^2$ & $-\tfrac1M \sum_i \log\!\left(\frac{\sum_j e^{-D_{ij}}}{M}\right)$ \\
  VFD \citep{romer2026save} & One reference member's path for each of the $K$ candidates; $V_\ell$ sums the across-member velocity variances (normalized by the number of members) over the chunk and averages over the scored candidates; $t_\ell \in \{0, 0.1, \ldots, 0.9, 0.95\}$ and $\Delta t_\ell = t_{\ell+1} - t_\ell$, with the last interval ending at $1$; members are the $M$ draws, or two trained policies for the ensemble reference & $\sum_\ell \Delta t_\ell\,\frac{t_\ell}{1-t_\ell}\,V_\ell$ \\
  \midrule
  CoMP-EKF (ours) & Recursion~\eqref{eq:comp_mean}--\eqref{eq:comp_state} with exact $J^{x}_i$ from batched reverse-mode autodiff; analytic $J^{\theta}_i$ and covariance products with $\Sigma_\theta$ & $U_{\mathrm{chunk}}$ or $U_{\mathrm{endpoint}}$ (Appendix~\ref{app:score}) \\
  CoMP-UKF (ours) & $2D+1$ state sigma points evaluated in batches through the action expert; regularized covariance solve for $A_i$ (Appendix~\ref{app:algo_ekf}); weight terms analytic in $J^{\theta}_i$ & $U_{\mathrm{chunk}}$ or $U_{\mathrm{endpoint}}$ (Appendix~\ref{app:score}) \\
  \bottomrule
  \end{tabular}
\end{table}

\FloatBarrier

\subsection{Supplementary Details and Results: Faithfulness of the Propagated Covariance}
\label{app:faithfulness}

For each task of \S\ref{sec:calibration}, we train a $29$M-parameter transformer variant of Diffusion Policy~\citep{chi2023diffusion} with a ResNet~\citep{he2016deep} image encoder on the $200$ proficient-human demonstrations, using the flow matching objective~\eqref{eq:cfm_loss}. Training runs for $200$ epochs with AdamW at learning rate $10^{-4}$ and batch size $2048$ for \textit{Lift} and $1024$ for the other tasks. The two-member ensembles of \S\ref{sec:failure} use the same training recipe. Each policy predicts $H = 16$ actions, integrated for $T = 50$ Euler steps.\looseness-1

The plain head is the Kronecker-factored Laplace of the last layer on the training split of each policy. Each policy uses a history of two observations, combining proprioception with external and wrist-camera images. The augmented head appends the two observation tokens formed from ResNet features and proprioception to condition the transformer, totaling $504$ units, to the $252$ last-layer features and is fit on the same windows. One robomimic policy trains for $200$ epochs in $8$ to $49$ minutes on an RTX 5090 depending on the task, whereas the Kronecker-factored Laplace fit over the same training windows takes under half a minute for either head.

\paragraph{Reference and estimators.}

The $150$ observation windows per task of \S\ref{sec:calibration} are drawn at random from the demonstration dataset. For each window, all estimators start from the same initial noise sample $x_0$. We construct the Gaussian reference using $M = 6000$ posterior draws integrated from $x_0$. BayesDiff uses $S = 24$ state particles and $M = 128$ posterior weight draws per step to estimate the velocity moments; its full-covariance ablation uses $S = 512$ particles per step. CoMP-EKF uses the exact state Jacobian and CoMP-UKF uses $2D + 1 = 225$ state sigma points, evaluated in a single batch through the transformer and its output layer.

\paragraph{Metrics.}

We compare empirical pairwise moments using the variogram discrepancy
\begin{equation}
\label{eq:variogram_discrepancy}
D_{\mathrm{VG}} \triangleq
\sum_{d,e=1}^{D}
\left[
\frac{1}{N}\sum_{n=1}^{N}|z_d^{(n)}-z_e^{(n)}|^{1/2}
-
\frac{1}{N}\sum_{n=1}^{N}|y_d^{(n)}-y_e^{(n)}|^{1/2}
\right]^2,
\end{equation}
\noindent
where $z^{(n)}$ are samples from the estimator's Gaussian and $y^{(n)}$ are held-out action chunks from the reference ensemble, with $N = 500$ for each. We report this discrepancy relative to that of the reference Gaussian on the same held-out chunks. The expert-action negative log-likelihood evaluates the ground-truth action chunk from each demonstration under each estimator's Gaussian; the ensemble reference for this metric uses $M = 1000$ posterior weight draws. Each bar of Figure~\ref{fig:calibration} is the median over the windows with a bootstrap $95\%$ interval of the median.

\paragraph{Additional results.}

\begin{figure}[!htpb]
  \centering
  \includegraphics[width=\textwidth]{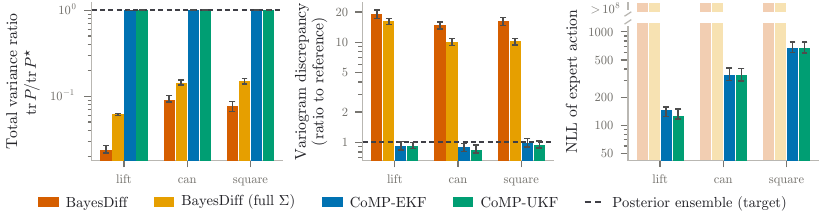}
  \caption{\textbf{Calibration with the augmented head on robomimic.} All estimators use the augmented last-layer Laplace posterior. Bars show medians over $150$ observation windows per task with bootstrap $95\%$ intervals; dashed lines mark reference values estimated from the augmented posterior. Both BayesDiff variants have expert-action NLLs above $10^8$.}
  \label{fig:calibration_aug}
\end{figure}

Figure~\ref{fig:calibration_aug} repeats the study of \S\ref{sec:calibration} with the augmented head for all estimators. Both CoMP variants remain close to the posterior reference in total variance and variogram discrepancy, while both BayesDiff variants still underestimate spread. Compared with the plain head (Figure~\ref{fig:calibration}), the augmented Laplace head lowers expert-action NLL by $8$ to $10\%$ for both CoMP variants on all three tasks.

\begin{figure}[!htpb]
  \centering
  \includegraphics[width=\textwidth]{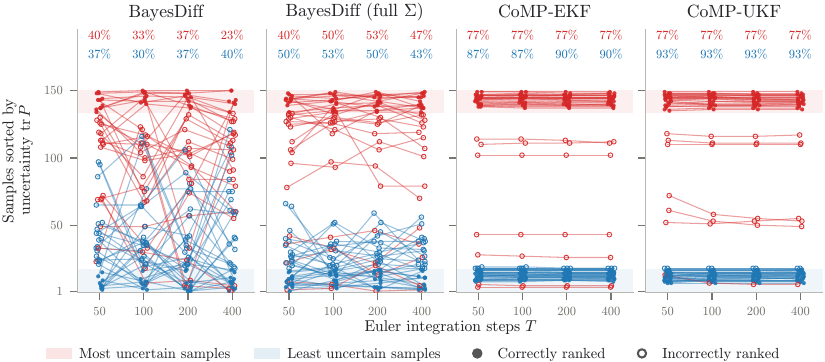}
  \caption{\rebuttal{\textbf{Ordering of the most and least uncertain observation windows on robomimic.} Each estimator scores the ten observations per task ranked most uncertain (red) and the ten ranked least uncertain (blue) by a $6000$-draw ensemble reference at $T = 400$ (i.e., $30$ observations per group over the three tasks). The shaded bands mark the estimator's own most and least uncertain observations, so a correctly ranked observation falls inside its band (filled) and an incorrectly ranked one outside it (hollow); percentages are the share of each group ranked correctly. BayesDiff mixes the two groups and reorders them across $T$, while CoMP keeps them apart and stable at every $T$.}}
  \label{fig:ranking_study}
\end{figure}

\rebuttal{The study of \S\ref{sec:calibration} shows the effect of variance collapse on the scale of the total variance, but not whether the collapse also changes which observations an estimator ranks as most and least uncertain. Figure~\ref{fig:ranking_study} tests this by repeating the experiment with the plain head at $T \in \{50, 100, 200, 400\}$ Euler steps, with the reference samples ranked using a $6000$-draw ensemble. We compute the reference's ranking at $T = 400$, where the integration error of the ensemble is smallest. For each task, we take the ten observations the reference ranks most uncertain and the ten it ranks least uncertain (thirty per group over the three tasks), and count at every $T$ how many of them the estimator also ranks among its most or least uncertain observations.} 

\rebuttal{BayesDiff (first panel) places at most $40\%$ of the observations correctly and reorders them from one step count to the next, with observations moving between its most and least uncertain groups across $T$; propagating the full covariance (second panel) raises the share to at most $53\%$ without repairing the ordering, in line with Corollary~\ref{cor:fullcov}. In contrast, both CoMP variants (third and fourth panels) keep the two groups apart at every $T$, ranking $77$ to $93\%$ of the observations correctly, and the few observations they misplace are the same ones at every $T$. Therefore, the number of integration steps not only collapses the scale of BayesDiff's total variance but also changes how it ranks observations, which explains why its downstream performance changes with the step count even under scale-invariant decision rules such as the conformal threshold~\eqref{eq:cp_threshold} and the selection rule~\eqref{eq:save_selection}. As the results with the augmented head closely resemble those of the plain head, they are not included.}\looseness-1

\clearpage

\subsection{Supplementary Details and Results: Response of the Augmented Head}
\label{app:robolab_tasks}

\paragraph{Policy and augmented head.}

We use openpi's $\pi_{0.5}$ fine-tuned on DROID with joint-position actions (release \texttt{pi05\_droid\_jointpos}). For $\pi_{0.5}$, we form $c(o)$ by separately averaging the image tokens of each of the two camera views and the language tokens of its PaliGemma vision-language backbone~\citep{beyer2024paligemma}. Appending these three $2048$-dimensional averages to the action expert's $1024$ last-layer features gives $7168$ features in total. The augmented head increases posterior storage from $4$ MB to $206$ MB, less than $3\%$ of $\pi_{0.5}$'s $7.2$ GB of weights.

\paragraph{Fitting and evaluation.}

For each of the ten tasks, we run $100$ episodes per wording with paired sampling seeds to measure task success. The policy predicts and executes a chunk of $H = 15$ actions per decision at $T = 10$ Euler steps. Each Laplace posterior is fit once for all ten tasks, using up to ten successful rollouts per task selected in order from episodes $0$ to $19$ of the ``specific'' wording. We use an equally weighted mixture of about $2.2$k observation frames from these RoboLab rollouts and $50$k DROID frames for the posterior fit. The conditioning features are standardized per dimension using DROID statistics.

\begin{figure}[!htpb]
  \centering
  \includegraphics[width=\textwidth]{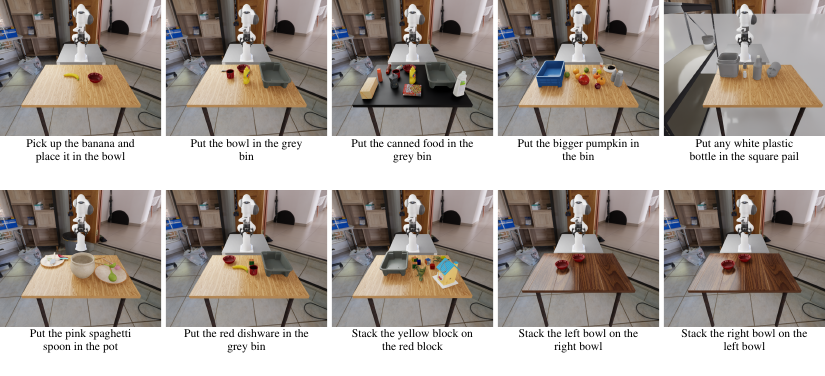}
  \caption{\textbf{The ten RoboLab tasks of \S\ref{sec:conditioning}.} Initial scene of each task from the front camera view, with its \emph{default} instruction; the study runs every task under three wordings (specific, default and vague) from the same scenes.}
  \label{fig:robolab_scenes}
\end{figure}

We evaluate uncertainty at the first decision of held-out episodes $20$ to $99$. The sampling estimators use $M = 16$ posterior weight draws, with VFD averaging $K=5$ action candidates. BayesDiff uses $S = 10$ state particles per step. Figure~\ref{fig:robolab_scenes} shows the ten scenes, and Table~\ref{tab:robolab_wordings} lists their instruction wordings.

\clearpage

\begin{table}[!htpb]
  \centering
  \caption{\textbf{Instruction wordings for the ten RoboLab tasks.} Each task is evaluated under specific, default and vague instructions from the same initial configuration in \S\ref{sec:conditioning}. Note how language instructions become more ambiguous from left to right.}
  \label{tab:robolab_wordings}
  \small
  \setlength{\tabcolsep}{5pt}
  \begin{tabular}{@{}>{\raggedright\arraybackslash}p{0.85in}>{\raggedright\arraybackslash}p{2.0in}>{\raggedright\arraybackslash}p{1.25in}>{\raggedright\arraybackslash}p{0.9in}@{}}
\toprule
\multirow{2}{*}{Task} & \multicolumn{3}{c}{Language instruction wording} \\
\cmidrule(l){2-4}
 & Specific & Default & Vague \\
\midrule
banana in bowl & Grasp the yellow banana and place it inside the red bowl on the table & Pick up the banana and place it in the bowl & Put the fruit in the bowl \\
\addlinespace[2pt]
pumpkin in bin & Identify the larger pumpkin of the two pumpkins on the table and place it in the bin & Put the bigger pumpkin in the bin & Put the big pumpkin away \\
\addlinespace[2pt]
bowl in bin & Pick up the bowl from the table and place it inside the grey bin & put the bowl in the grey bin & put away bowl \\
\addlinespace[2pt]
stack L on R & Pick up the bowl to the left of the robot and place it on top of the bowl to the right of the robot & Stack the left bowl on the right bowl & Stack the left bowl on the right \\
\addlinespace[2pt]
stack R on L & Pick up the bowl to the right of the robot and place it on top of the bowl to the left of the robot & Stack the right bowl on the left bowl & Stack the right bowl on the left \\
\addlinespace[2pt]
canned food in bin & Put the small canned tuna in the grey bin & Put the canned food in the grey bin & Put away cans \\
\addlinespace[2pt]
bottle in pail & Pick up one of the three white plastic bottles and place it inside the square pail & Put any white plastic bottle in the square pail & Put a bottle in the square pail \\
\addlinespace[2pt]
spoon in pot & Pick up the pink spaghetti serving spoon that's on the left plate and place it inside the pot in the center of the table & Put the pink spaghetti spoon in the pot & Put the pink spoon in the pot \\
\addlinespace[2pt]
red dishes in bin & Pick up each red mug and bowl from the table and place them in the grey bin & Put the red dishware in the grey bin & Clean up red dishware \\
\addlinespace[2pt]
stack yellow & Pick up the yellow block on the left and place it precisely on top of the red block in front of the birdhouse & Stack the yellow block on the red block & Stack blocks: yellow on red \\
\bottomrule
\end{tabular}

\end{table}

\paragraph{Additional results.}

\begin{figure}[!htpb]
  \centering
  \includegraphics[width=\textwidth]{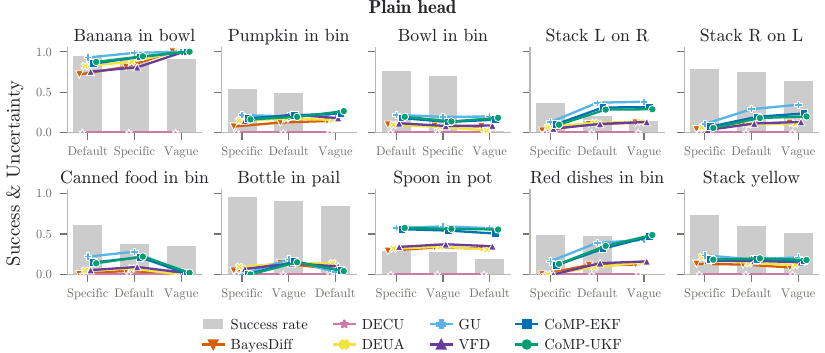}\\[8pt]
  \includegraphics[width=\textwidth]{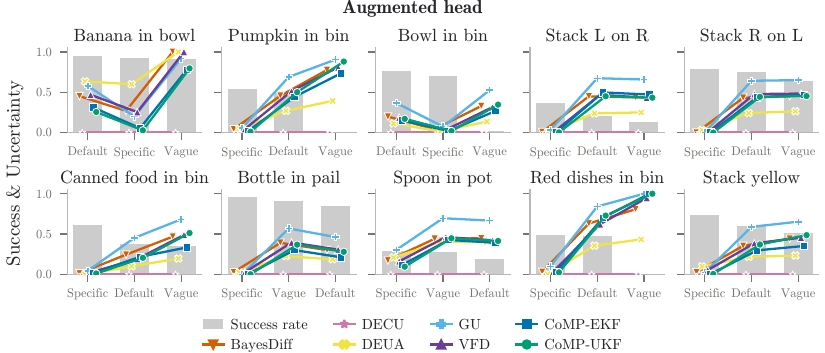}
  \caption{\textbf{RoboLab instruction shift, per task.} Success rate (grey) and first-decision uncertainty (lines, min-max scaled to $[0,1]$ per estimator and Laplace fit across tasks and wordings) for the ten tasks and three wordings, under the plain (top) and augmented (bottom) Laplace heads. With the augmented head, every estimator except DECU generally reports higher uncertainty as language instructions become more ambiguous, while the plain head remains mostly flat, highlighting the benefits of our augmented Laplace posterior.}
  \label{fig:robolab_shift_tasks}
\end{figure}

Figure~\ref{fig:robolab_shift_tasks} shows the per-task results underlying Figure~\ref{fig:robolab_shift}, for both the plain and augmented Laplace posteriors. With the plain head, uncertainty estimates remain mostly flat even as language instructions become more ambiguous. In contrast, the augmented head strengthens the response to these changes for every estimator except DECU.

\begin{figure}[!htpb]
  \centering
  \includegraphics[width=\textwidth]{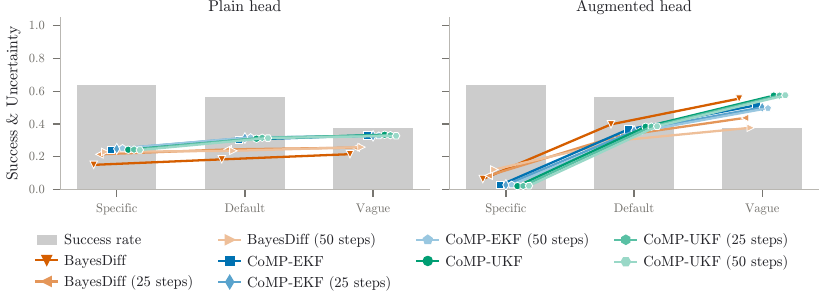}
  \caption{\textbf{BayesDiff's response weakens with more integration steps.} Success rate (grey) and first-decision uncertainty, averaged over the ten RoboLab tasks by wording, for BayesDiff, CoMP-EKF and CoMP-UKF at $10$ Euler integration steps (as in Figure~\ref{fig:robolab_shift}), $25$ and $50$. On both Laplace heads, each CoMP variant's response remains nearly unchanged as the number of integration steps increases, while BayesDiff's response to instruction ambiguity becomes flatter.}
  \label{fig:robolab_shift_steps}
\end{figure}

Although the augmented head makes BayesDiff uncertainty estimates change with language ambiguity, it does not resolve the variance collapse issue outlined in \S\ref{sec:comp}. Figure~\ref{fig:robolab_shift_steps} compares BayesDiff, CoMP-EKF and CoMP-UKF at $10$, $25$ and $50$ Euler integration steps on the same recorded decisions. As the step count grows, BayesDiff's uncertainty shrinks and its response to instruction ambiguity weakens, while both CoMP variants retain their sensitivity. The default $10$-step sampler of $\pi_{0.5}$ therefore makes this limitation less apparent, but increasing the number of integration steps reveals it even with the augmented head.

\FloatBarrier

\subsection{Supplementary Details and Results: Runtime Failure Detection}
\label{app:fiper_tables}

\paragraph{Robomimic.}

We use the policies and Laplace posteriors of Appendix~\ref{app:faithfulness}. We evaluate six perturbation magnitudes and an unperturbed condition, with $50$ episodes per condition for a total of $350$ test rollouts per task. On \textit{Lift} and \textit{NutAssemblySquare}, we displace both the target object in the table plane and the end effector from its initial pose along random directions. The respective magnitude ranges are $0.056$--$0.116$ m and $0.050$--$0.120$ m for \textit{Lift}, and $0.005$--$0.045$ m and $0.026$--$0.095$ m for \textit{NutAssemblySquare}. On \textit{PickPlaceCan}, we perturb only the end effector by $0.057$--$0.202$ m and leave the can's position unperturbed; its initial position already varies across the demonstrations.

\begin{figure}[!htpb]
  \centering
  \includegraphics[width=\textwidth]{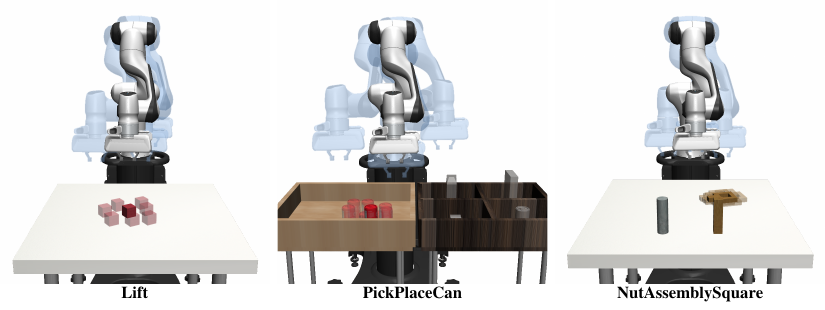}
  \caption{\textbf{Perturbations on robomimic.} The nominal initial state of each task from the front camera (solid), with perturbed resets overlaid as ghosts: the target object displaced in the table plane (translucent) and the end effector displaced from its reset pose (blue). Walls and floor are removed for clarity. The object displacements on PickPlaceCan are shown for illustration only, as its sweep perturbs the end effector alone.}
  \label{fig:robomimic_perturb}
\end{figure}

An episode runs for at most $400$ environment steps with an execution horizon of $h = 8$ and $T = 50$ Euler integration steps, giving at most $E = 50$ decisions. Thresholds are calibrated on a separate set of $20$ successful, unperturbed episodes per task. BayesDiff estimates its cross term with $S = 24$ state particles per step, the sampling estimators draw $M = 16$ posterior weight samples, and VFD averages the scores of $K=5$ action candidates. For comparison and following~\cite{romer2026save}, we also report DECU and VFD with ensembles of two independently trained policies, following the training recipe of Appendix~\ref{app:faithfulness}. Results for the sampling estimators are averaged over five random seeds. Figure~\ref{fig:robomimic_perturb} illustrates the perturbations on the three robomimic tasks.

\paragraph{LIBERO-90.}

\begin{figure}[!htpb]
  \centering
  \includegraphics[width=\textwidth]{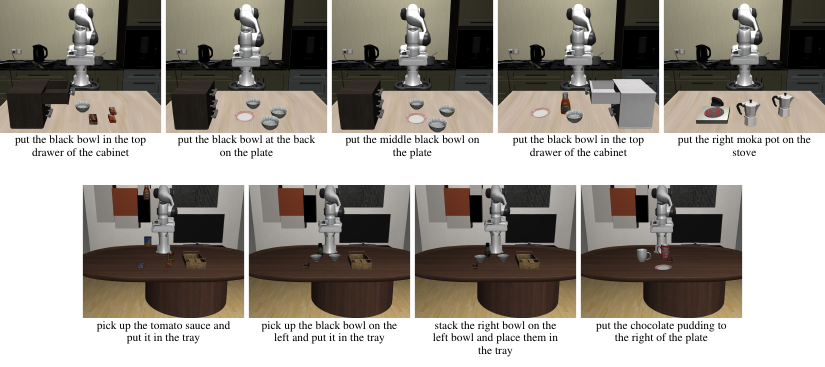}
  \caption{\textbf{The nine LIBERO-90 tasks of \S\ref{sec:failure}.} Initial scene and instruction of each task from the simulator's front camera.}
  \label{fig:libero90_scenes}
\end{figure}

We use openpi's \texttt{pi05\_libero} checkpoint of $\pi_{0.5}$, fine-tuned on the LIBERO-Spatial, Object, Goal and 10 suites~\citep{liu2023libero}. We select LIBERO-90 tasks $2$, $12$, $14$, $29$, $38$, $59$, $60$, $64$ and $70$, for which the policy's success rate lies between $0.3$ and $0.7$, following the evaluation criterion in FIPER~\citep{romer2025fiper}. Figure~\ref{fig:libero90_scenes} shows their initial scenes and instructions. Both Laplace posteriors are fit once using $30$k observations from the four training suites; the augmented head appends the same pooled image and language features as in Appendix~\ref{app:robolab_tasks}, standardized per dimension using these observations. Each Laplace fit takes $21$ minutes on one A100.

We run $50$ episodes per task, each capped at $400$ environment steps. The policy predicts chunks of $H = 10$ actions with $T = 10$ Euler integration steps and executes $h = 5$ actions per decision, giving at most $E = 80$ decisions. Thresholds are calibrated on $20$ successful episodes per task, drawn from two calibration runs on scenes disjoint from the test set. BayesDiff estimates its cross term with $S = 10$ state particles per step, the sampling estimators draw $M = 16$ posterior weight samples, and VFD averages the scores of $K=5$ action candidates. Results for the sampling estimators are averaged over five random seeds.

\paragraph{Threshold calibration and evaluation.}

Following FIPER~\citep{romer2025fiper}, we aggregate uncertainty over a trailing window of $w = 5$ decisions to reduce sensitivity to isolated score fluctuations. Let $U_t$ denote the resulting windowed score and $\tau \triangleq (o_t, a^{(t)}_{1:H})_{t=1}^{E_\tau}$ a rollout of $E_\tau$ decisions, with observation $o_t$ and predicted action chunk $a^{(t)}_{1:H}$ at decision $t$. For each estimator and task, thresholds are fitted on a calibration set $\mathcal{D}_{\mathrm{c}} = \{\tau^i\}_{i=1}^{N}$ of $N = 20$ successful rollouts, disjoint from the test episodes. We use FIPER's constant conformal prediction threshold, computed as the empirical $q$-quantile of the per-rollout maxima:
\begin{equation}
\label{eq:cp_threshold}
\gamma_q \triangleq \mathrm{Quantile}_q\!\left(\left\{\max_{t} U^i_t\right\}_{i=1}^{N}\right),
\qquad
F(\tau_{:t}) \triangleq \mathbf{1}\!\left[U_t > \gamma_q\right],
\end{equation}
where $U^i_t$ is the windowed score at decision $t$ of calibration rollout $\tau^i$, and $F(\tau_{:t}) \in \{0,1\}$ is the binary failure alarm based on the test rollout up to decision $t$. A test rollout is flagged at the first decision for which $F(\tau_{:t}) = 1$, with the threshold $\gamma_q$ held fixed throughout the episode. We use the same window size and threshold rule for every estimator and average the evaluation metrics over ten quantiles $q \in \{0.90, 0.91, \dots, 0.99\}$ so results are not cherry picked. Results for the sampling estimators are further averaged over five random seeds.

\paragraph{Metrics.}

Throughout the failure detection experiments, we use balanced accuracy, $\mathrm{Acc} \triangleq (\mathrm{TPR} + \mathrm{TNR})/2$, following FIPER~\citep{romer2025fiper}. The true positive rate (TPR) is the fraction of failed rollouts correctly flagged, and the true negative rate (TNR) is the fraction of successful rollouts left unflagged. For a correctly flagged failure, normalized detection time is $\mathrm{DT} \triangleq (t_{\mathrm{det}} - 1)/(N_{\mathrm{steps}} - 1)$, where $t_{\mathrm{det}}$ is the policy step at which the first alarm is raised and $N_{\mathrm{steps}}$ is the maximum number of policy steps in a test episode for that task. The timestep-wise accuracy (TWA) discounts late failure detections as
\begin{equation}
\label{eq:fiper_twa}
\mathrm{TWA} \triangleq \frac{1}{2}\left[\mathrm{TNR} + \mathrm{TPR}\,(1-\mathrm{DT})\right],
\end{equation}
where $\mathrm{DT}$ is the mean normalized detection time over correctly flagged failures. All estimators score the same rollouts recorded with the MAP policy, so they receive identical observations and differences in detection arise from their uncertainty scores only.

\paragraph{Additional results.}

Table~\ref{tab:fiper_heads} compares all estimators with both Laplace heads on robomimic and LIBERO-90, allowing the effect of the augmented Laplace posterior to be assessed for each method. \rebuttal{On robomimic, our augmented head improves accuracy and detection time for five of the seven post-hoc estimators, and TWA for BayesDiff, DECU and CoMP-EKF, while GU and VFD do not benefit from it.} \rebuttal{On LIBERO-90, it raises CoMP-UKF's TWA from $0.631$ to $0.643$ and lowers DT from $0.514$ to $0.474$.} With either head, both CoMP variants outperform BayesDiff on all three suite-averaged metrics, while CoMP-UKF remains competitive with the strongest post-hoc baselines.

\clearpage

\begin{table}[!htb]
\centering
\caption{\textbf{Runtime failure detection on robomimic and LIBERO-90 with both Laplace heads.} Timestep-wise accuracy, balanced accuracy and normalized detection time using the conformal threshold~\eqref{eq:cp_threshold} with $w = 5$, as mean $\pm$ standard deviation over five sampling seeds. Rows show individual robomimic tasks, their average, and the average over nine LIBERO-90 tasks. All post-hoc estimators use the plain head (top) or our augmented head (bottom). The best post-hoc value in each row is bold and the second best underlined. Trained ensembles are references evaluated only on robomimic.}
\label{tab:fiper_heads}
\scriptsize
\vspace{3mm}
\setlength{\tabcolsep}{1.8pt}
\resizebox{\textwidth}{!}{\rebuttal{\begin{tabular}{llccccccccc}
\toprule
 & & \multicolumn{7}{c}{Post-hoc, single model} & \multicolumn{2}{c}{Trained ensembles} \\
\cmidrule(lr){3-9} \cmidrule(lr){10-11}
Task & Metric & BayesDiff & DECU & DEUA & GU & VFD & CoMP-EKF & CoMP-UKF & DECU-ens & VFD-ens \\
 & &  &  &  &  &  & (ours) & (ours) &  &  \\
\midrule
\multicolumn{11}{l}{\emph{Plain head}} \\
\midrule
\multirow{3}{*}{can}
 & TWA $\uparrow$ & 0.602{\tiny$\pm$0.049} & 0.721{\tiny$\pm$0.003} & 0.731{\tiny$\pm$0.006} & 0.729{\tiny$\pm$0.004} & \textbf{0.754{\tiny$\pm$0.001}} & 0.722{\tiny$\pm$0.032} & \underline{0.753{\tiny$\pm$0.002}} & 0.777{\tiny$\pm$0.006} & 0.762{\tiny$\pm$0.002} \\
 & Acc. $\uparrow$ & 0.677{\tiny$\pm$0.075} & 0.807{\tiny$\pm$0.004} & 0.813{\tiny$\pm$0.006} & 0.784{\tiny$\pm$0.005} & \underline{0.838{\tiny$\pm$0.001}} & 0.811{\tiny$\pm$0.038} & \textbf{0.839{\tiny$\pm$0.003}} & 0.886{\tiny$\pm$0.011} & 0.862{\tiny$\pm$0.001} \\
 & DT $\downarrow$ & 0.355{\tiny$\pm$0.057} & 0.260{\tiny$\pm$0.007} & 0.239{\tiny$\pm$0.011} & \textbf{0.138{\tiny$\pm$0.007}} & \underline{0.224{\tiny$\pm$0.003}} & 0.271{\tiny$\pm$0.038} & 0.233{\tiny$\pm$0.011} & 0.226{\tiny$\pm$0.012} & 0.244{\tiny$\pm$0.004} \\
\midrule
\multirow{3}{*}{lift}
 & TWA $\uparrow$ & 0.556{\tiny$\pm$0.022} & \textbf{0.631{\tiny$\pm$0.002}} & \underline{0.630{\tiny$\pm$0.003}} & 0.618{\tiny$\pm$0.005} & 0.629{\tiny$\pm$0.001} & 0.604{\tiny$\pm$0.032} & 0.621{\tiny$\pm$0.001} & 0.654{\tiny$\pm$0.003} & 0.646{\tiny$\pm$0.001} \\
 & Acc. $\uparrow$ & 0.682{\tiny$\pm$0.039} & \textbf{0.791{\tiny$\pm$0.003}} & \underline{0.789{\tiny$\pm$0.006}} & 0.772{\tiny$\pm$0.006} & 0.780{\tiny$\pm$0.003} & 0.745{\tiny$\pm$0.069} & 0.773{\tiny$\pm$0.001} & 0.825{\tiny$\pm$0.003} & 0.808{\tiny$\pm$0.001} \\
 & DT $\downarrow$ & 0.589{\tiny$\pm$0.079} & 0.408{\tiny$\pm$0.006} & 0.406{\tiny$\pm$0.013} & \underline{0.392{\tiny$\pm$0.005}} & \textbf{0.384{\tiny$\pm$0.005}} & 0.462{\tiny$\pm$0.060} & 0.393{\tiny$\pm$0.002} & 0.430{\tiny$\pm$0.004} & 0.435{\tiny$\pm$0.002} \\
\midrule
\multirow{3}{*}{square}
 & TWA $\uparrow$ & 0.503{\tiny$\pm$0.013} & 0.542{\tiny$\pm$0.003} & 0.549{\tiny$\pm$0.008} & \underline{0.568{\tiny$\pm$0.008}} & 0.554{\tiny$\pm$0.002} & 0.550{\tiny$\pm$0.023} & \textbf{0.569{\tiny$\pm$0.005}} & 0.591{\tiny$\pm$0.008} & 0.654{\tiny$\pm$0.004} \\
 & Acc. $\uparrow$ & 0.598{\tiny$\pm$0.041} & 0.657{\tiny$\pm$0.006} & 0.670{\tiny$\pm$0.015} & 0.659{\tiny$\pm$0.011} & \textbf{0.698{\tiny$\pm$0.007}} & 0.662{\tiny$\pm$0.056} & \underline{0.698{\tiny$\pm$0.007}} & 0.762{\tiny$\pm$0.007} & 0.822{\tiny$\pm$0.003} \\
 & DT $\downarrow$ & 0.507{\tiny$\pm$0.076} & 0.467{\tiny$\pm$0.007} & 0.459{\tiny$\pm$0.035} & \textbf{0.430{\tiny$\pm$0.036}} & 0.451{\tiny$\pm$0.005} & 0.503{\tiny$\pm$0.065} & \underline{0.435{\tiny$\pm$0.011}} & 0.467{\tiny$\pm$0.023} & 0.414{\tiny$\pm$0.006} \\
\midrule
\multirow{3}{*}{Average}
 & TWA $\uparrow$ & 0.554{\tiny$\pm$0.021} & 0.631{\tiny$\pm$0.001} & 0.637{\tiny$\pm$0.004} & 0.638{\tiny$\pm$0.003} & \underline{0.646{\tiny$\pm$0.001}} & 0.625{\tiny$\pm$0.016} & \textbf{0.648{\tiny$\pm$0.002}} & 0.674{\tiny$\pm$0.004} & 0.687{\tiny$\pm$0.002} \\
 & Acc. $\uparrow$ & 0.652{\tiny$\pm$0.036} & 0.752{\tiny$\pm$0.002} & 0.757{\tiny$\pm$0.007} & 0.738{\tiny$\pm$0.003} & \textbf{0.772{\tiny$\pm$0.002}} & 0.739{\tiny$\pm$0.028} & \underline{0.770{\tiny$\pm$0.002}} & 0.825{\tiny$\pm$0.004} & 0.831{\tiny$\pm$0.001} \\
 & DT $\downarrow$ & 0.484{\tiny$\pm$0.045} & 0.378{\tiny$\pm$0.006} & 0.368{\tiny$\pm$0.009} & \textbf{0.320{\tiny$\pm$0.014}} & \underline{0.353{\tiny$\pm$0.002}} & 0.412{\tiny$\pm$0.032} & 0.354{\tiny$\pm$0.004} & 0.374{\tiny$\pm$0.010} & 0.365{\tiny$\pm$0.003} \\
\midrule
\multirow{3}{*}{LIBERO-90}
 & TWA $\uparrow$ & 0.533{\tiny$\pm$0.013} & 0.518{\tiny$\pm$0.011} & 0.596{\tiny$\pm$0.004} & 0.626{\tiny$\pm$0.004} & \textbf{0.635{\tiny$\pm$0.006}} & 0.584{\tiny$\pm$0.012} & \underline{0.631{\tiny$\pm$0.011}} & -- & -- \\
 & Acc. $\uparrow$ & 0.633{\tiny$\pm$0.023} & 0.591{\tiny$\pm$0.015} & 0.688{\tiny$\pm$0.010} & \underline{0.770{\tiny$\pm$0.003}} & 0.757{\tiny$\pm$0.008} & 0.685{\tiny$\pm$0.018} & \textbf{0.783{\tiny$\pm$0.013}} & -- & -- \\
 & DT $\downarrow$ & 0.587{\tiny$\pm$0.025} & 0.578{\tiny$\pm$0.029} & 0.592{\tiny$\pm$0.048} & 0.526{\tiny$\pm$0.027} & \textbf{0.504{\tiny$\pm$0.029}} & 0.548{\tiny$\pm$0.019} & \underline{0.514{\tiny$\pm$0.017}} & -- & -- \\
\midrule
\midrule
\multicolumn{11}{l}{\emph{Augmented head}} \\
\midrule
\multirow{3}{*}{can}
 & TWA $\uparrow$ & 0.671{\tiny$\pm$0.026} & \textbf{0.735{\tiny$\pm$0.005}} & 0.723{\tiny$\pm$0.005} & 0.691{\tiny$\pm$0.005} & 0.722{\tiny$\pm$0.002} & \underline{0.728{\tiny$\pm$0.007}} & 0.725{\tiny$\pm$0.001} & 0.777{\tiny$\pm$0.006} & 0.762{\tiny$\pm$0.002} \\
 & Acc. $\uparrow$ & \underline{0.803{\tiny$\pm$0.017}} & \textbf{0.806{\tiny$\pm$0.006}} & 0.794{\tiny$\pm$0.007} & 0.722{\tiny$\pm$0.007} & 0.792{\tiny$\pm$0.003} & 0.798{\tiny$\pm$0.010} & 0.793{\tiny$\pm$0.001} & 0.886{\tiny$\pm$0.011} & 0.862{\tiny$\pm$0.001} \\
 & DT $\downarrow$ & 0.386{\tiny$\pm$0.071} & 0.180{\tiny$\pm$0.004} & 0.179{\tiny$\pm$0.004} & \textbf{0.073{\tiny$\pm$0.004}} & 0.175{\tiny$\pm$0.003} & 0.177{\tiny$\pm$0.008} & \underline{0.172{\tiny$\pm$0.001}} & 0.226{\tiny$\pm$0.012} & 0.244{\tiny$\pm$0.004} \\
\midrule
\multirow{3}{*}{lift}
 & TWA $\uparrow$ & 0.616{\tiny$\pm$0.001} & \textbf{0.652{\tiny$\pm$0.008}} & 0.621{\tiny$\pm$0.001} & \underline{0.633{\tiny$\pm$0.001}} & 0.620{\tiny$\pm$0.001} & 0.620{\tiny$\pm$0.002} & 0.620{\tiny$\pm$0.000} & 0.654{\tiny$\pm$0.003} & 0.646{\tiny$\pm$0.001} \\
 & Acc. $\uparrow$ & 0.776{\tiny$\pm$0.001} & \textbf{0.810{\tiny$\pm$0.011}} & 0.787{\tiny$\pm$0.001} & \underline{0.801{\tiny$\pm$0.001}} & 0.785{\tiny$\pm$0.001} & 0.782{\tiny$\pm$0.005} & 0.784{\tiny$\pm$0.000} & 0.825{\tiny$\pm$0.003} & 0.808{\tiny$\pm$0.001} \\
 & DT $\downarrow$ & 0.429{\tiny$\pm$0.003} & \textbf{0.381{\tiny$\pm$0.011}} & 0.435{\tiny$\pm$0.003} & \underline{0.426{\tiny$\pm$0.001}} & 0.436{\tiny$\pm$0.001} & 0.438{\tiny$\pm$0.016} & 0.431{\tiny$\pm$0.000} & 0.430{\tiny$\pm$0.004} & 0.435{\tiny$\pm$0.002} \\
\midrule
\multirow{3}{*}{square}
 & TWA $\uparrow$ & 0.513{\tiny$\pm$0.015} & 0.548{\tiny$\pm$0.004} & 0.568{\tiny$\pm$0.006} & 0.550{\tiny$\pm$0.007} & 0.566{\tiny$\pm$0.003} & \underline{0.569{\tiny$\pm$0.025}} & \textbf{0.599{\tiny$\pm$0.002}} & 0.591{\tiny$\pm$0.008} & 0.654{\tiny$\pm$0.004} \\
 & Acc. $\uparrow$ & 0.641{\tiny$\pm$0.036} & 0.710{\tiny$\pm$0.006} & \underline{0.731{\tiny$\pm$0.006}} & 0.657{\tiny$\pm$0.008} & 0.730{\tiny$\pm$0.004} & 0.701{\tiny$\pm$0.041} & \textbf{0.740{\tiny$\pm$0.003}} & 0.762{\tiny$\pm$0.007} & 0.822{\tiny$\pm$0.003} \\
 & DT $\downarrow$ & 0.547{\tiny$\pm$0.041} & 0.453{\tiny$\pm$0.017} & \underline{0.443{\tiny$\pm$0.021}} & 0.478{\tiny$\pm$0.020} & 0.449{\tiny$\pm$0.019} & 0.467{\tiny$\pm$0.088} & \textbf{0.383{\tiny$\pm$0.006}} & 0.467{\tiny$\pm$0.023} & 0.414{\tiny$\pm$0.006} \\
\midrule
\multirow{3}{*}{Average}
 & TWA $\uparrow$ & 0.600{\tiny$\pm$0.011} & \underline{0.645{\tiny$\pm$0.002}} & 0.637{\tiny$\pm$0.003} & 0.625{\tiny$\pm$0.003} & 0.636{\tiny$\pm$0.001} & 0.639{\tiny$\pm$0.007} & \textbf{0.648{\tiny$\pm$0.001}} & 0.674{\tiny$\pm$0.004} & 0.687{\tiny$\pm$0.002} \\
 & Acc. $\uparrow$ & 0.740{\tiny$\pm$0.015} & \textbf{0.775{\tiny$\pm$0.004}} & 0.771{\tiny$\pm$0.003} & 0.727{\tiny$\pm$0.003} & 0.769{\tiny$\pm$0.001} & 0.760{\tiny$\pm$0.012} & \underline{0.773{\tiny$\pm$0.001}} & 0.825{\tiny$\pm$0.004} & 0.831{\tiny$\pm$0.001} \\
 & DT $\downarrow$ & 0.454{\tiny$\pm$0.023} & 0.338{\tiny$\pm$0.007} & 0.352{\tiny$\pm$0.006} & \textbf{0.325{\tiny$\pm$0.006}} & 0.353{\tiny$\pm$0.007} & 0.361{\tiny$\pm$0.030} & \underline{0.329{\tiny$\pm$0.002}} & 0.374{\tiny$\pm$0.010} & 0.365{\tiny$\pm$0.003} \\
\midrule
\multirow{3}{*}{LIBERO-90}
 & TWA $\uparrow$ & 0.548{\tiny$\pm$0.012} & 0.522{\tiny$\pm$0.008} & 0.574{\tiny$\pm$0.009} & \underline{0.609{\tiny$\pm$0.009}} & 0.533{\tiny$\pm$0.012} & 0.575{\tiny$\pm$0.015} & \textbf{0.643{\tiny$\pm$0.005}} & -- & -- \\
 & Acc. $\uparrow$ & 0.664{\tiny$\pm$0.012} & 0.605{\tiny$\pm$0.008} & 0.665{\tiny$\pm$0.007} & \underline{0.741{\tiny$\pm$0.018}} & 0.613{\tiny$\pm$0.011} & 0.679{\tiny$\pm$0.021} & \textbf{0.794{\tiny$\pm$0.005}} & -- & -- \\
 & DT $\downarrow$ & 0.608{\tiny$\pm$0.027} & 0.609{\tiny$\pm$0.041} & 0.581{\tiny$\pm$0.046} & \underline{0.499{\tiny$\pm$0.022}} & 0.525{\tiny$\pm$0.036} & 0.557{\tiny$\pm$0.028} & \textbf{0.474{\tiny$\pm$0.014}} & -- & -- \\
\bottomrule
\end{tabular}
}}
\end{table}

\begin{table}[!htb]
\centering
\caption{\rebuttal{\textbf{Compute-matched comparison with a posterior ensemble.} An ensemble of size $M = 2D + 1$ drawn from the augmented Laplace posterior and scored with $U_{\mathrm{endpoint}}$, against CoMP-EKF and CoMP-UKF. Metrics and protocol follow Table~\ref{tab:fiper_heads}, averaged over tasks, as mean $\pm$ standard deviation over five sampling seeds. Evals.\ gives the number of model evaluations per Euler step; for CoMP-EKF these are backward passes. The best value in each column is bold.}}
\label{tab:sampling_budget}
\scriptsize
\vspace{3mm}
\setlength{\tabcolsep}{3pt}
{\color{rebuttalblue}\begin{tabular}{@{}lcccccccc@{}}
\toprule
 & \multicolumn{4}{c}{robomimic} & \multicolumn{4}{c}{LIBERO-90} \\
\cmidrule(lr){2-5} \cmidrule(lr){6-9}
Estimator & Evals. & TWA $\uparrow$ & Acc. $\uparrow$ & DT $\downarrow$ & Evals. & TWA $\uparrow$ & Acc. $\uparrow$ & DT $\downarrow$ \\
\midrule
Posterior ensemble, $U_{\mathrm{endpoint}}$ & 225 & \textbf{0.648}{\tiny$\pm$0.002} & 0.772{\tiny$\pm$0.002} & 0.332{\tiny$\pm$0.003} & 141 & 0.611{\tiny$\pm$0.005} & 0.763{\tiny$\pm$0.011} & 0.529{\tiny$\pm$0.010} \\
CoMP-EKF (ours) & 112 & 0.639{\tiny$\pm$0.007} & 0.760{\tiny$\pm$0.012} & 0.361{\tiny$\pm$0.030} & 70 & 0.575{\tiny$\pm$0.015} & 0.679{\tiny$\pm$0.021} & 0.557{\tiny$\pm$0.028} \\
CoMP-UKF (ours) & 225 & \textbf{0.648}{\tiny$\pm$0.001} & \textbf{0.773}{\tiny$\pm$0.001} & \textbf{0.329}{\tiny$\pm$0.002} & 141 & \textbf{0.643}{\tiny$\pm$0.005} & \textbf{0.794}{\tiny$\pm$0.005} & \textbf{0.474}{\tiny$\pm$0.014} \\
\bottomrule
\end{tabular}
}
\end{table}

\paragraph{\rebuttal{Compute-matched comparison with a posterior ensemble.}}

\rebuttal{We also compare CoMP against an ensemble of $M$ heads drawn from the augmented Laplace posterior, each integrated through the flow and scored with $U_{\mathrm{endpoint}}$ in~\eqref{eq:score} from the sample covariance of their chunks. With $M = 2D + 1$, that is, $225$ draws on robomimic and $141$ on LIBERO-90, the ensemble performs the same number of model evaluations per Euler step as CoMP-UKF. Table~\ref{tab:sampling_budget} shows that, at equal budget, the ensemble is on par with CoMP-UKF on robomimic, whereas on LIBERO-90 CoMP-UKF leads across all metrics. In contrast, CoMP-EKF trails both, which we attribute to the linearization error around the MAP path accumulating over the Euler steps. Overall, CoMP-UKF matches or surpasses a posterior ensemble at the same computational budget.}

\subsection{Supplementary Details and Results: Uncertainty-Guided Fine-Tuning}
\label{app:oneshot}

\paragraph{Policy and posterior fitting.}

We use SmolVLA with the SmolVLM2-500M backbone and an action expert with $480$-dimensional last-layer features, following SAVE's training recipe~\citep{romer2026save}. The policy takes two $256 \times 256$ images, proprioception and a language instruction as input, predicts $H = 50$ actions and executes $h = 25$. We train it for $30$k steps at batch size $32$ on the LIBERO-Spatial, Object and Goal suites and LIBERO-10 tasks $0$ to $2$. Each policy occupies $1.2$ GB (about $0.6$B parameters), with a LIBERO-10 success rate of $20$ to $25\%$ before fine-tuning. We train ten independent policies, two for each of five evaluation seeds. All post-hoc estimators use one policy per seed, while the two-member VFD baseline~\citep{romer2026save} uses both to score $K = 5$ candidate chunks.

For the plain Laplace posterior, we fit a Kronecker-factored Laplace posterior over the expert's output layer with $480$ input features, using $30$k frames stratified over the training episodes. The augmented head appends $3840$ features obtained by averaging the backbone's token features separately for each camera view, the language instruction and the proprioceptive state. This increases posterior storage from $0.95$ MB to $75$ MB. One SmolVLA member trains for $30$k steps in $6.5$ GPU-hours on a single A100, and each $30$k-frame head fit takes $1.8$ GPU-hours, of which the forward pass of the frozen policy is the bulk.

\begin{figure}[!htpb]
  \centering
  \includegraphics[width=\textwidth]{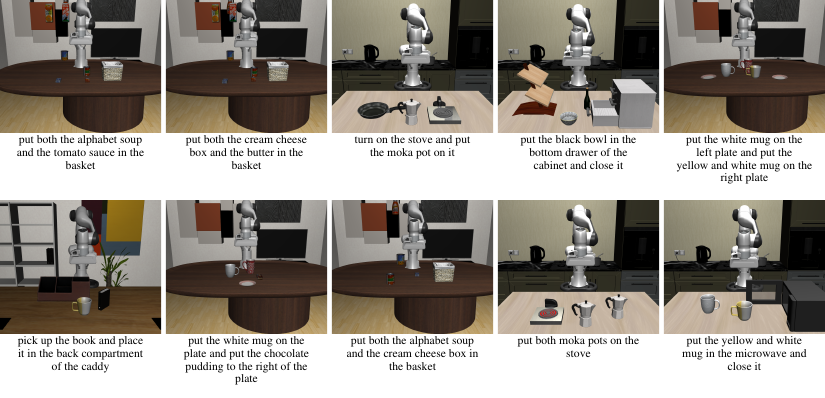}
  \caption{\textbf{The ten LIBERO-10 tasks of \S\ref{sec:finetuning}.} Initial scene and instruction of each task from the simulator's front camera.}
  \label{fig:libero10_scenes}
\end{figure}

\paragraph{Demonstration selection.}

The candidate pool contains all LIBERO-10 demonstrations from the HuggingFaceVLA release\footnote{\url{https://huggingface.co/datasets/HuggingFaceVLA/libero}}. Each demonstration is scored once at its first frame with $T = 10$, $K = 8$ candidates for Laplace-based VFD, $M = 16$ posterior draws for the sampling estimators and $S = 24$ particles for BayesDiff. We select demonstrations once per run for budgets $X \in \{25, 50, 75, 100, 150, 200, 250\}$. Following SAVE~\citep{romer2026save}, we compute task sampling probabilities as
\begin{equation}
\label{eq:save_selection}
\bar U_j \triangleq \frac{1}{|\mathcal{D}_j|}\sum_{d \in \mathcal{D}_j} U(d),
\qquad
p(j) = \frac{\bar U_j^{2.5}}{\sum_k \bar U_k^{2.5}},
\end{equation}
where $\mathcal{D}_j$ is the demonstration pool for task $j$ and $U(d)$ is the uncertainty score at the first frame of demonstration $d$. We sample tasks according to $p(j)$ and select their highest-scoring demonstrations until the budget $X$ is met, using uniform task sampling if all task scores are zero.

\paragraph{Fine-tuning and evaluation.}

For each seed, all fine-tuning runs start from the same pre-trained checkpoint. We use AdamW for $8000$ steps at batch size $32$, sampling equally from the selected demonstrations and the original training data. The learning rate warms up to $5 \times 10^{-5}$ over $200$ steps, followed by cosine decay to $5 \times 10^{-6}$. Each fine-tuned policy is evaluated on all ten LIBERO-10 tasks (Figure~\ref{fig:libero10_scenes}).

\paragraph{Metrics and additional results.}

We assess demonstration selection with two metrics. The first measures the demonstrations needed to reach $90\%$ of random selection's mean success over the five seeds at $X = 250$, using linear interpolation of each policy's success curve. The second measures the average improvement over random selection across budgets. For each method and seed, we compute the weighted average:
\begin{equation}
\label{eq:oneshot_gainz}
\mathrm{Gain} \triangleq \sum_X w_X\bigl[\mathrm{SR}(X)-\mathrm{SR}_{\mathrm{rand}}(X)\bigr],
\end{equation}

where $\mathrm{SR}(X)$ and $\mathrm{SR}_{\mathrm{rand}}(X)$ are the method's and random selection's success rates for the same seed, measured in percent and averaged over all ten LIBERO-10 tasks. The weights $w_X$ are the normalized trapezoidal weights, accounting for the spacing between budgets and summing to one. Gain is therefore measured in percentage points. Both metrics are reported as means and standard deviations over the five trained policies.

\begin{figure}[!htpb]
  \centering
  \includegraphics[width=\textwidth]{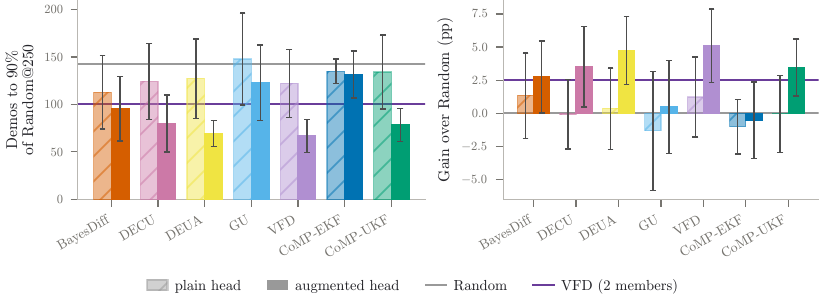}
  \caption{\textbf{Effect of augmented Laplace head on data selection.} Demonstrations needed to reach $90\%$ of random selection's mean success at $X = 250$ (left; lower is better) and mean gain over random across budgets (right; higher is better), on all ten LIBERO-10 tasks. Paired bars compare plain and augmented heads for every estimator, with mean and standard deviation over five pre-trained policies. Horizontal lines mark random selection and two-member VFD. Overall, our augmented Laplace head improves both metrics for every estimator, with CoMP-UKF and several baselines exceeding the two-member VFD ensemble's mean gain using only one policy.}
  \label{fig:oneshot_heads}
\end{figure}

\begin{figure}[htpb!]
  \centering
  \includegraphics[width=\textwidth]{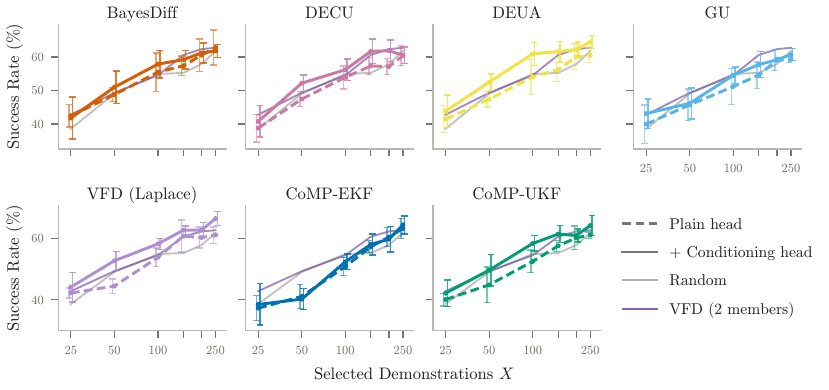}
  \caption{\textbf{Data selection on LIBERO-10, per estimator.} Success rate after fine-tuning on the ten LIBERO-10 tasks against the number of selected demonstrations $X$, for every estimator on the plain head (dashed) and on the augmented head (solid), with random selection and the two-member VFD as references; mean and standard deviation over five pre-trained policies. On the plain Laplace head every estimator tracks random selection; our augmented head brings most estimators above random selection, with several also surpassing the two-member VFD.} 
  \label{fig:oneshot_curves}
\end{figure}

Figure~\ref{fig:oneshot_heads} extends Figure~\ref{fig:oneshot_gain} to compare plain and augmented Laplace heads for every estimator; Figure~\ref{fig:oneshot_curves} gives the full success curves. Our augmented Laplace head increases mean gain and reduces the average number of demonstrations needed for every estimator by reusing observation embeddings the policy already computes. With a single policy, CoMP-UKF and several augmented baselines exceed the two-member VFD ensemble's mean gain, while CoMP-EKF remains close to random selection. Although BayesDiff benefits from the augmented head at $10$ integration steps, the following ablation shows that this gain disappears as the step count increases, consistent with its variance collapse.

\paragraph{Integration steps.}

Consistent with the variance decay theorem in~\S\ref{sec:comp}, BayesDiff's fine-tuning advantage over random selection disappears as more integration steps are used to score demonstrations (Table~\ref{tab:oneshot_steps}). We compare $10$, $25$ and $50$ Euler integration steps, keeping the fine-tuning protocol fixed and always evaluating the resulting policies at $10$ steps. Even with our augmented Laplace posterior, BayesDiff's average gain nearly vanishes at $25$ steps and turns negative at $50$, with the same declining trend under the plain Laplace posterior. In contrast, CoMP-UKF retains its gain with the augmented head across all three step counts. CoMP-EKF's gains are less consistent across step counts and repetitions, which may reflect its first-order linearization; CoMP-UKF instead propagates sigma points through the nonlinear velocity field avoiding this issue.

\begin{table}[!htpb]
\centering
\caption{\textbf{BayesDiff's data selection weakens as integration steps increase.} Average gain over random selection (pp) and demonstrations needed to reach $90\%$ of random selection's mean success at $X = 250$ on the ten LIBERO-10 tasks. Selection scores use $10$, $25$ or $50$ Euler integration steps; the fine-tuning protocol is fixed and evaluation always uses $10$ integration steps. Mean and standard deviation over five pre-trained policies, with the plain (top) and augmented Laplace head (bottom). The best value in each row is bold and the second best underlined. BayesDiff's gain declines as integration steps increase, while CoMP-UKF retains its gain with our augmented head.}\label{tab:oneshot_steps}
\small
\setlength{\tabcolsep}{5pt}
\renewcommand{\arraystretch}{0.95}  
\vspace{5mm}
\begin{tabular}{llccc}
\toprule
Metric & Steps & BayesDiff & CoMP-EKF & CoMP-UKF \\
 & & & (ours) & (ours) \\
\midrule
\multicolumn{5}{l}{\emph{Plain head}} \\
\midrule
\multirow{3}{*}{Gain over random (pp) $\uparrow$}
 & 10 & \textbf{+1.3 $\pm$ 3.2} & -1.0 $\pm$ 2.1 & \underline{0.0 $\pm$ 2.9} \\
 & 25 & -2.6 $\pm$ 4.3 & \underline{-0.5 $\pm$ 3.8} & \textbf{+0.4 $\pm$ 3.2} \\
 & 50 & -3.8 $\pm$ 5.4 & \underline{-0.6 $\pm$ 2.9} & \textbf{+1.1 $\pm$ 2.8} \\
\midrule
\multirow{3}{*}{Demos to $90\%$ of random $\downarrow$}
 & 10 & \textbf{113 $\pm$ 43} & 135 $\pm$ 15 & \underline{134 $\pm$ 44} \\
 & 25 & 148 $\pm$ 58 & \underline{130 $\pm$ 33} & \textbf{126 $\pm$ 52} \\
 & 50 & 157 $\pm$ 45 & \underline{142 $\pm$ 33} & \textbf{119 $\pm$ 30} \\
\midrule\midrule
\multicolumn{5}{l}{\emph{Augmented head}} \\
\midrule
\multirow{3}{*}{Gain over random (pp) $\uparrow$}
 & 10 & \underline{+2.7 $\pm$ 2.7} & -0.5 $\pm$ 2.9 & \textbf{+3.5 $\pm$ 2.1} \\
 & 25 & +0.5 $\pm$ 2.5 & \underline{+2.5 $\pm$ 3.2} & \textbf{+3.3 $\pm$ 2.2} \\
 & 50 & -2.9 $\pm$ 3.8 & \underline{+1.7 $\pm$ 4.1} & \textbf{+3.7 $\pm$ 2.5} \\
\midrule
\multirow{3}{*}{Demos to $90\%$ of random $\downarrow$}
 & 10 & \underline{95 $\pm$ 38} & 132 $\pm$ 28 & \textbf{78 $\pm$ 19} \\
 & 25 & 118 $\pm$ 44 & \underline{102 $\pm$ 20} & \textbf{74 $\pm$ 16} \\
 & 50 & 150 $\pm$ 31 & \underline{108 $\pm$ 41} & \textbf{84 $\pm$ 7} \\
\bottomrule
\end{tabular}

\end{table}

\end{document}